\documentclass{article}

\usepackage[numbers]{natbib}
 \usepackage[preprint]{neurips_2026}

\usepackage[utf8]{inputenc} 
\usepackage[T1]{fontenc}    
\usepackage{hyperref}       
\usepackage{url}            
\usepackage{booktabs}       
\usepackage{amsfonts}       
\usepackage{nicefrac}       
\usepackage{microtype}      
\usepackage{xcolor}         

\usepackage{subcaption}
\usepackage{bbding}
\usepackage{multirow}
\usepackage{booktabs}
\usepackage{graphicx}
\usepackage{amsmath}
\usepackage{enumitem}

\title{Semantic-Driven Scale and Spatial Selection for Efficient Cross-Modal Alignment in Referring Remote Sensing Image Segmentation}

\author{
\textbf{Kun Li}$^{1, 3}$
\quad
\textbf{Shengxi Gui}$^{2}$
\quad
\textbf{Francesco Nex}$^{3}$
\quad
\textbf{Michael Ying Yang}$^{4}$
\\
$^1$ University of Liverpool, UK
\quad
$^2$ University of Michigan, USA
\\
$^3$ University of Twente, NL
\quad
$^4$ University of Bath, UK
}

\begin{document}

\maketitle

\begin{abstract}
  Referring Remote Sensing Image Segmentation (RRSIS) seeks to localize and segment the target object or region specified by a natural language expression in a remote sensing image, requiring joint understanding of visual and linguistic information.
While existing RRSIS models have benefited from large-scale foundation models (\textit{e.g.}, CLIP), they predominantly rely on full fine-tuning.
These approaches are computationally intensive and may weaken the generalization ability of pre-trained models, as extensive fine-tuning on significantly smaller downstream datasets can distort the well-structured feature representations learned during large-scale pre-training.
Although Parameter-Efficient Tuning (PET) offers a potential alternative, existing PET frameworks primarily focus on single-modal optimization, failing to capture the complex cross-modal dependencies required for multimodal reasoning, while simultaneously struggling to bridge the substantial domain gap between natural scenes and aerial imagery.
To address these limitations, we propose a novel framework, Semantic-driven Scale and Spatial Selection for Efficient Cross-modal Alignment (S$^4$ECA), which enables effective and efficient cross-modal interaction through parameter-efficient adaptation.
Specifically, we design a dual-encoder adapter architecture.
The textual adapter employs learnable queries to distill highly semantic language proxies from word-level embeddings, facilitating early grounding.
Simultaneously, the visual adapter refines hierarchical feature representations through a multi-scale dense extractor, followed by a language-guided scale and spatial selection mechanism that dynamically emphasizes relevant visual contexts, ensuring precise cross-modal alignment.
By updating only 2.4\% of the backbone parameters, our proposed model achieves state-of-the-art performance on the RRSIS-D and RefSegRS datasets, demonstrating superior efficiency and precision in complex aerial scenarios.
\end{abstract}

\section{Introduction}
\label{sec: intro}
Referring Remote Sensing Image Segmentation (RRSIS) aims to identify and segment the target region within an aerial image based on a natural language expression.
In contrast to traditional vision-only segmentation tasks with fixed semantic categories, RRSIS requires fine-grained cross-modal reasoning to associate open-vocabulary textual expressions with the corresponding visual regions.
Beyond its theoretical significance, RRSIS has significant potential for advancing human-centered geospatial intelligence by enabling intuitive language-based interaction with remote sensing imagery.
Such capability is valuable for a wide range of applications, including urban monitoring \cite{duan2016towardsurban, li2024hrvqa}, disaster response \cite{rahnemoonfar2021floodnet, kalluri2024robustdisaster}, precision agriculture \cite{upadhyay2025deepagriculture}, and land-use change survey \cite{liu2024remoteclipchange}, where users often need to identify specific targets through natural language descriptions rather than predefined semantic categories.
However, accurate pixel-level localization remains highly challenging due to the substantial variations in object scale, complex spatial relationships among targets, and the cluttered backgrounds in high-resolution aerial imagery.

\begin{figure}[t]
  \centering
  \begin{subfigure}[t]{0.42\linewidth}
  \includegraphics[width=1\linewidth]{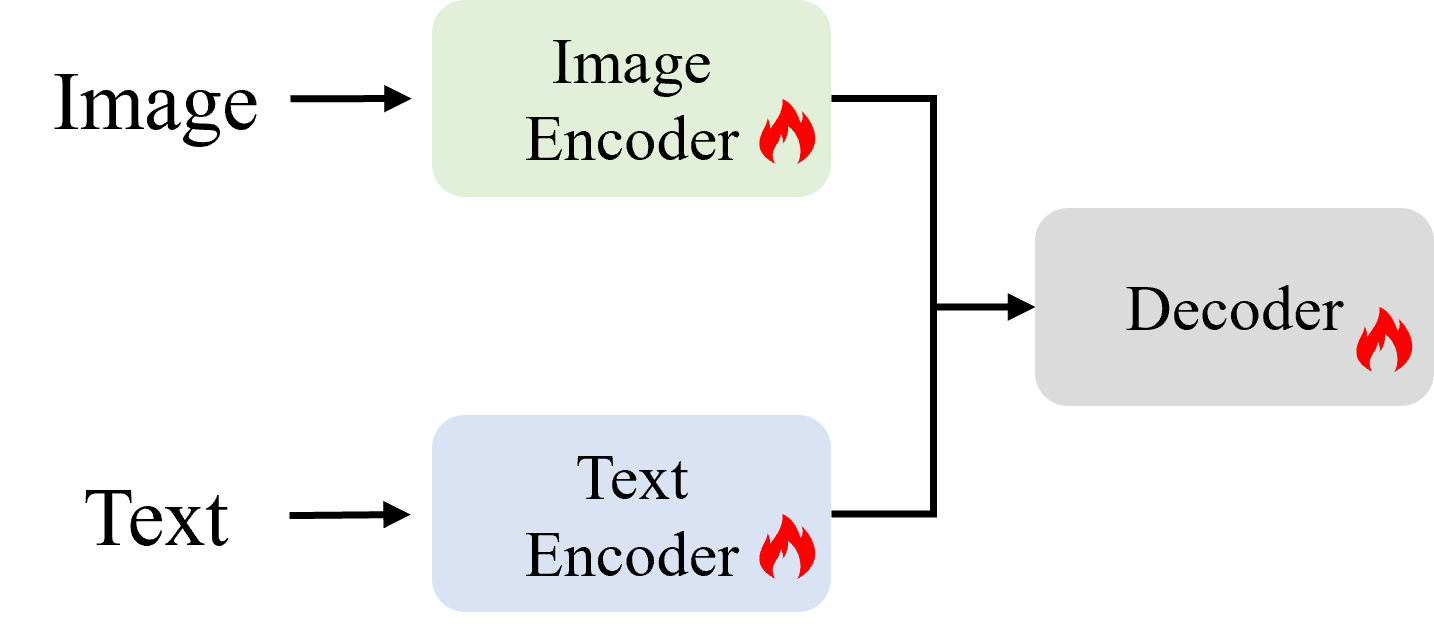}
  \subcaption{Full fine-tuning.}
  \label{fig: 1a}
  \end{subfigure}
  \hfill
  \begin{subfigure}[t]{0.42\linewidth}
  \includegraphics[width=1\linewidth]{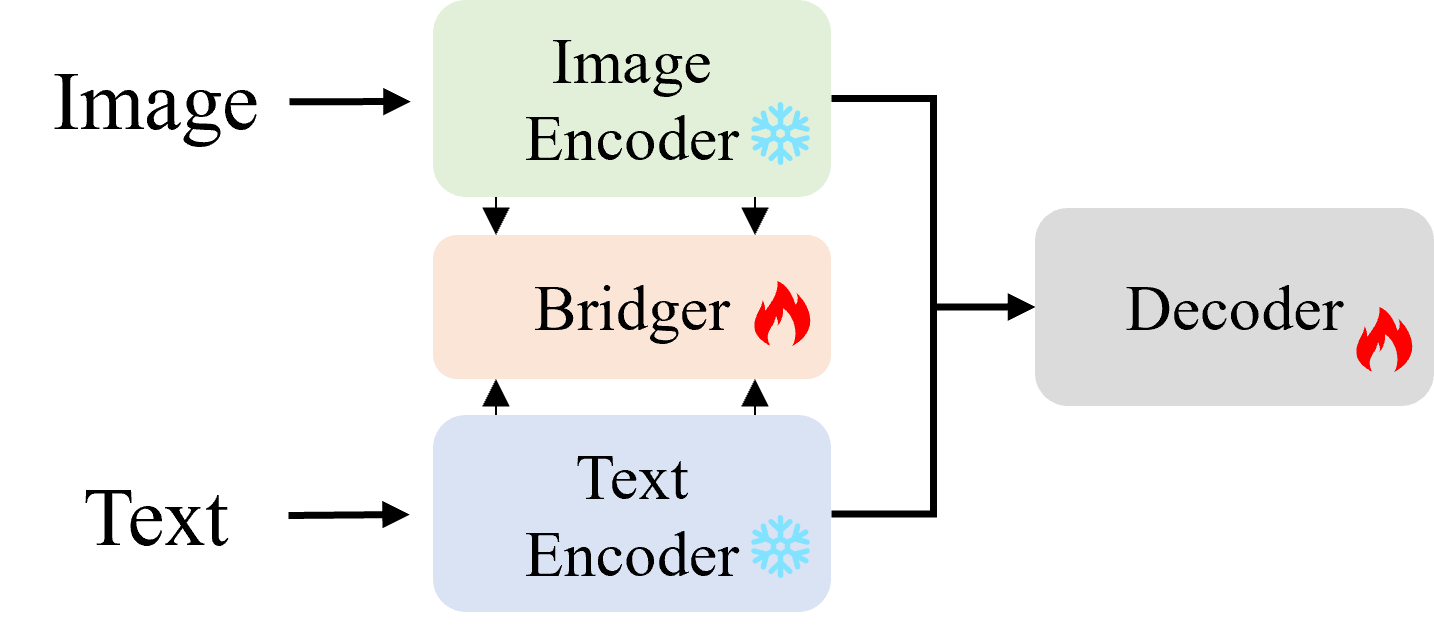}
  \subcaption{Bridger-based PET.}
  \label{fig: 1b}
  \end{subfigure}

  \begin{subfigure}[t]{0.42\linewidth}
  \includegraphics[width=1\linewidth]{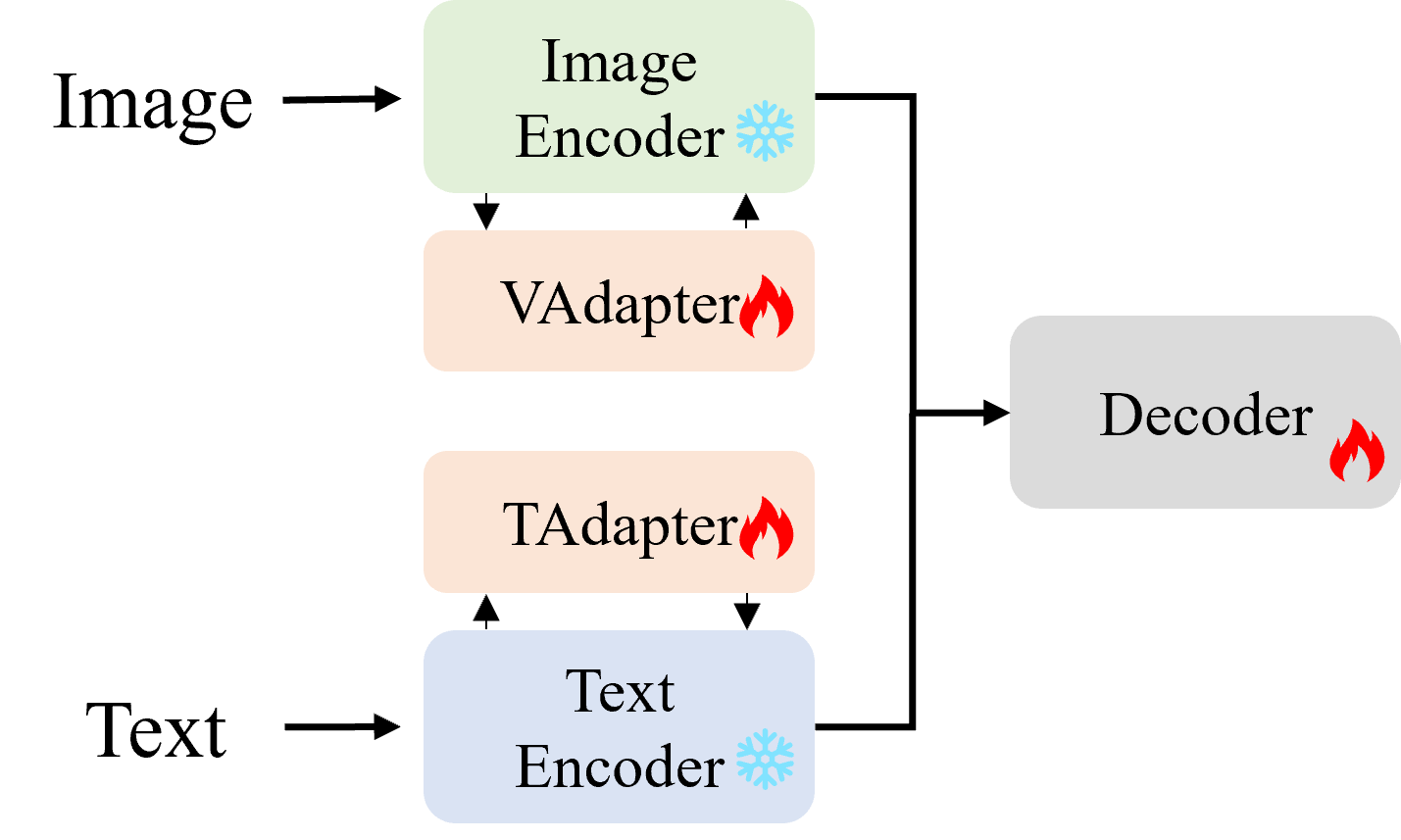}
  \subcaption{Dual-adapter-based PET.}
  \label{fig: 1c}
  \end{subfigure}
  \hfill
  \begin{subfigure}[t]{0.42\linewidth}
  \includegraphics[width=1\linewidth]{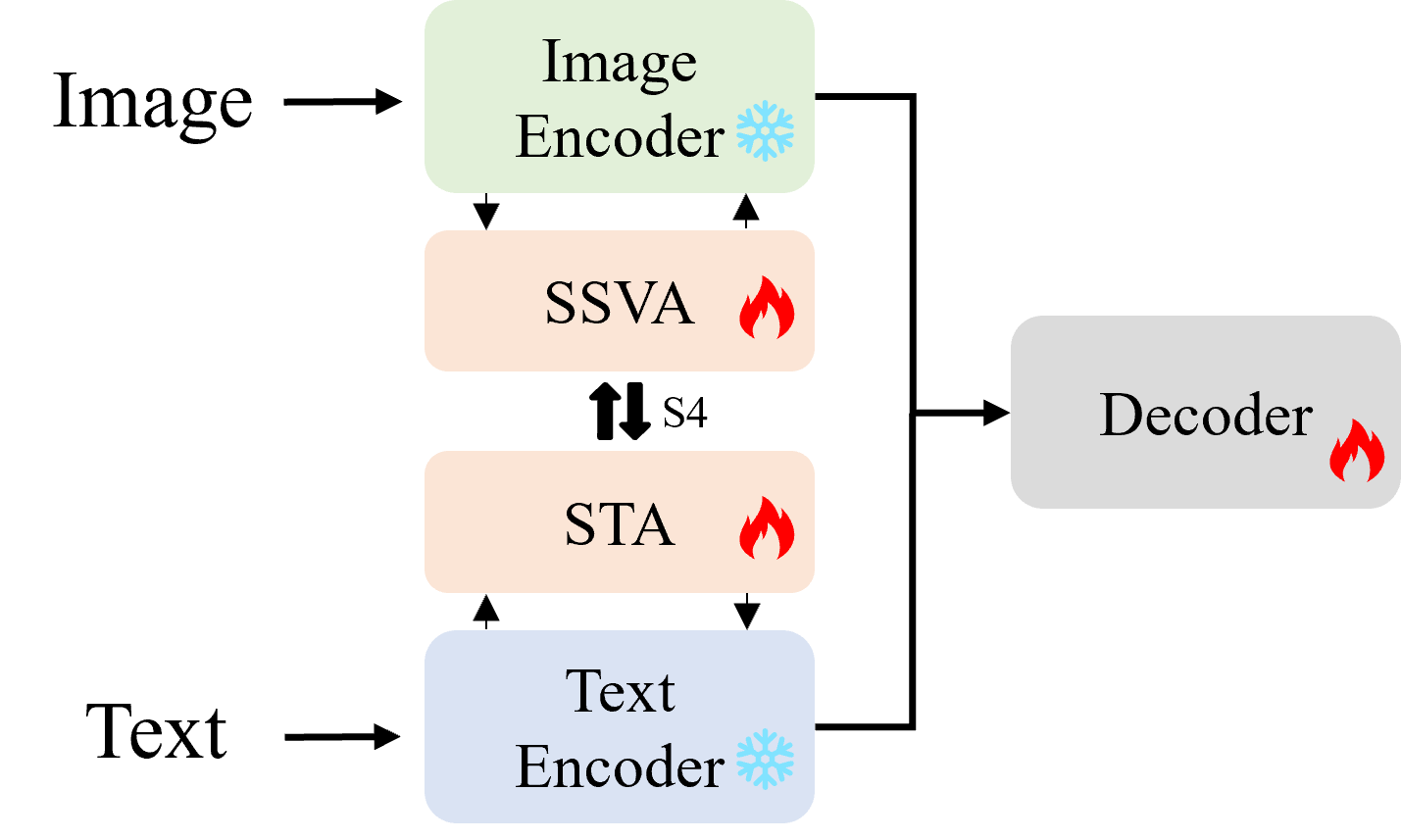}
  \subcaption{Our S4 adapter-based PET.}
  \label{fig: 1d}
  \end{subfigure}
  
  \caption{Illustration of distinct paradigms for the referring image segmentation task.
  (a) Traditional full fine-tuning methods update the entire pre-trained vision-language model on the target task.
  (b) Standard adapter-based parameter-efficient tuning (PET) methods introduce lightweight bottleneck modules (Bridger) into the frozen backbones to reduce parameter overhead.
 (c) Dual-adapter-based PET methods employ both visual and textual adapters, facilitating the efficient transfer via modular updates.
 (d) Our proposed S$^4$ECA method builds upon the dual-adapter architecture with our novel semantic-driven scale and spatial selection (S4) strategy by updating only a small set of tunable parameters.
  }
  \label{fig: illustration}
\end{figure}

Benefiting from recent advances in vision-language learning \cite{radford2021clip, yang2022lavt, liu2023caris, li2026qstar}, RRSIS has witnessed rapid progress in multimodal representation learning and cross-modal interaction.
Existing methods, such as LGCE \cite{yuan2024rrsis} and RMSIN \cite{liu2024rmsin}, generally follow a standard representation-fusion-segmentation paradigm, where hierarchical visual features are progressively integrated with linguistic representations extracted by pre-trained language encoders to facilitate target localization and segmentation.
Despite their promising performance, existing RRSIS frameworks primarily rely on full-parameter fine-tuning for domain adaptation (as shown in Figure~\ref{fig: 1a}).
Such a paradigm faces two fundamental challenges.
First, optimizing all model parameters on remote sensing datasets that are substantially smaller than the original pre-training corpus may compromise the transferability of pre-trained representations, limiting the preservation of general knowledge.
Second, updating the entire parameter space introduces considerable computational and memory overhead, reducing scalability and limiting practical deployment in real-world applications.

To address these challenges, Parameter-Efficient Tuning (PET) has emerged as a promising alternative for adapting large-scale foundation models to downstream tasks \cite{houlsby2019parameter1, chen2024convpet}.
Instead of updating the entire model, PET freezes most pre-trained parameters and introduces a small set of task-specific trainable components, enabling efficient adaptation while largely preserving the rich knowledge acquired during pre-training.
As a result, PET substantially reduces computational and memory costs without sacrificing the representational capacity of pre-trained models.
However, existing PET techniques are predominantly applied to single-modality tasks (\textit{e.g.},~image classification) and remain inadequately explored in the context of RRSIS.
Although a few pioneering studies have extended PET to dense prediction tasks, they primarily focus on intra-modal adaptation, while overlooking the complex cross-modal reasoning required to accurately align free-form textual descriptions with fine-grained visual content.
Consequently, they struggle to capture the complementary dependencies between language and vision that are essential for precise pixel-level localization.
Moreover, these methods are typically developed on natural image domains and therefore exhibit limited adaptability to remote sensing scenarios, where large-scale spatial variation, dense object distribution, and highly cluttered backgrounds introduce significantly greater complexity.
For instance, existing bridger-based designs such as ETRIS \cite{xu2023etris} do not explicitly model multi-scale visual structures, which are crucial for handling objects that vary dramatically in size within aerial imagery.
In addition, current methods place limited emphasis on textual adaptation, despite the fact that reliable segmentation in complex remote sensing scenes requires precise linguistic grounding to disambiguate visually similar targets.
For instance, DETRIS \cite{huang2025densely} relies solely on intra-modal updates within its textual adapter design, without explicitly modeling the dynamic interaction between textual representations and evolving visual features.
Consequently, these limitations significantly restrict their capacity to achieve robust cross-modal alignment and precise segmentation under the complex and diverse conditions of aerial imagery.

In this paper, we propose the \textbf{S}emantic-driven \textbf{S}cale and \textbf{S}patial \textbf{S}election based \textbf{E}fficient \textbf{C}ross-modal \textbf{A}lignment network (S$^4$ECA), an effective and efficient RRSIS framework designed for domain-specific adaptation.
The framework adopts a dual-adapter architecture that performs modality-specific adaptation for both visual and textual encoders while preserving the majority of pre-trained parameters.
Different from existing PET methods that apply static or intra-modal layer-wise tuning (as shown in Figure~\ref{fig: illustration}), our proposed textual adapter introduces a set of learnable queries to represent high-level semantic language proxies for early grounding.
Specifically, a lightweight language enhancer is first employed to capture local and global dependencies among textual tokens, after which the learnable queries aggregate complementary semantic information from the linguistic features.
These semantic proxies are subsequently interacted with visual representations through a lightweight cross-attention mechanism, enabling early cross-modal interaction.
By jointly modeling local textual structures and global semantic cues, the proposed design enhances language understanding while achieving efficient textual adaptation with substantially lower computational cost than conventional dense cross-attention.
To effectively adapt visual representations to the complex characteristics of remote sensing imagery, a dedicated convolutional module is further designed to extract multi-scale visual features.
Nevertheless, visual features alone often contain substantial redundancy and ambiguity in aerial scenes, where visually similar regions may correspond to entirely different semantic concepts.
To address this issue, a semantic-driven scale and spatial selection module is introduced to progressively refine visual representations under linguistic guidance.
The scale selection strategy first exploits the textual semantics to identify the most relevant feature scales for the queried target, while the spatial selection strategy further leverages the language proxies to emphasize informative regions and suppress irrelevant responses.
Through the joint modeling of semantic, scale, and spatial information, the proposed visual adapter produces more discriminative and task-relevant representations for cross-modal alignment.
Finally, the adapted visual and linguistic features are integrated through an effective fusion neck and decoded by a standard segmentation head to generate the target mask.
Extensive experiments on the RRSIS-D and RefSegRS benchmarks demonstrate that our proposed S$^4$ECA achieves state-of-the-art performance, while updating only 2.4\% of the backbone parameters.
Further ablation studies validate the effectiveness of our parameter-efficient designs, demonstrating its capacity to preserve transferable pre-trained representations while successfully adapting to the unique challenges of remote sensing image interpretation.


The \textbf{main contributions} of this paper are summarized as follows:
\begin{itemize}[leftmargin=*, itemsep=0pt]
    \item We propose \textbf{S$^4$ECA}, a novel parameter-efficient tuning framework that adapts large-scale foundation models to the RRSIS domain.
    By decoupling the adaptation process from conventional full fine-tuning, our framework preserves the structural integrity of pre-trained feature spaces while enabling robust cross-modal alignment for dense prediction in complex aerial imagery.
    \item We introduce a lightweight textual adapter to distill high-level semantic proxies from word-level embeddings through a set of learnable language queries.
    This mechanism effectively captures local textual dependencies and facilitates early grounding, improving the discrimination of referred targets.
    \item We design a dedicated visual adapter structured to enhance hierarchical feature representations through a multi-scale dense extractor.
    This module is further coupled with a semantic-guided scale and spatial selection mechanism, which dynamically suppresses redundant background noise and promotes the adaptation of discriminative visual features.
    \item Extensive experiments on two challenging RRSIS benchmarks demonstrate that our S$^4$ECA achieves state-of-the-art segmentation performance with only 2.4\% tunable parameters of the backbone.
\end{itemize}

\section{Related Work}
\label{related work}

\subsection{Referring Remote Sensing Image Segmentation}
Referring Image Segmentation (RIS) aims to localize and segment target objects within images based on natural language expressions.
Early efforts \cite{nagaraja2016modelingearly2, li2018rrn} employ simple concatenation to fuse visual and linguistic features for this language-guided image segmentation task.
The recent adoption of Transformer-based \cite{vaswani2017attention} architectures has further accelerated progress, enabling robust cross-modal integration through attention and language-aware feature alignment \cite{ding2021visionvlt, liu2023gres}.
Despite these advancements, standard RIS methods (\textit{e.g.},~LAVT \cite{yang2022lavt}, CRIS \cite{wang2022cris}, and CARIS \cite{liu2023caris}) often struggle when adapted to high-resolution aerial imagery, where target objects exhibit extreme scale variations and are embedded in complex, cluttered backgrounds.
To bridge this domain gap, the emerging field of Referring Remote Sensing Image Segmentation (RRSIS) has sought to address these unique aerial challenges.
Pioneering works introduced dedicated datasets and benchmarks, such as RefSegRS \cite{yuan2024rrsis} and the larger-scale RRSIS-D \cite{liu2024rmsin}, which have facilitated the evaluation of mainstream RIS architectures in aerial contexts.
DANet \cite{pan2024mmrrsis} attempted to optimize performance through explicit affinity alignment and alleviate the impact of deceptive noise interference with reliable agent alignment.
FIANet \cite{lei2024exploringarxiv} enhanced cross-modal alignment through fine-grained image-text interactions by decoupling referring expressions into object and spatial descriptions, while further improving multi-scale feature fusion with text-aware self-attention mechanisms.
However, these methods still treat linguistic features as fixed representations from the original language encoder and equally apply uniform attention mechanisms to all textual tokens during alignment, which often neglects the spatial relationships in both visual contexts and textual descriptions.
To overcome these limitations, SBANet \cite{li2025sbanet} jointly updated visual and linguistic features during bidirectional cross-modal alignment and introduced a text-conditioned channel-spatial aggregator to enhance multi-scale feature interaction and spatial reasoning.
CroBIM-U \cite{sun2026crobim-u} proposed an uncertainty-guided framework that leverages a pixel-wise referring uncertainty map to enable adaptive language-vision fusion and region-aware refinement for improving disambiguation in aerial scenarios.
RS2SAM2 \cite{rong2026rs2sam2} adapted the SAM2 \cite{ravi2025sam2} model for this task by introducing a bidirectional hierarchical framework that incorporates pseudo-mask-based dense prompting and enforces boundary-aware constraints to enhance segmentation accuracy.

Despite these advancements, existing RRSIS methods \cite{liu2024rmsin, ma2025lscf, li2025sbanet, jia2025egat, zhang2026htvg} often rely on full fine-tuning, which presents significant drawbacks.
Large-scale pre-trained vision models (\textit{e.g.},~CLIP \cite{radford2021clip}, Swin Transformer \cite{liu2021swin}, SAM \cite{kirillov2023sam}) provide a rich foundation of generalizable features, but this knowledge is inherently fragile.
The aggressive optimization characteristic of full fine-tuning tends to distort these latent representations, risking the loss of pre-trained feature hierarchies in favor of narrow, domain-specific patterns.
This degradation is further exacerbated by the substantial domain gap between natural images and the complex, high-resolution aerial images.
Consequently, these models often suffer from suboptimal generalization.
To address these challenges, our work shifts toward a parameter-efficient tuning strategy, which preserves the structural integrity of the pre-trained backbone while injecting lightweight, learnable modules that adapt the model to the nuanced requirements of RRSIS with significantly reduced computational overhead.

\subsection{Parameter-Efficient Tuning}
Parameter-Efficient Tuning (PET) has emerged as a compelling paradigm for transfer learning, enabling the adaptation of large-scale foundation models by freezing primary backbones and updating only a minimal subset of parameters \cite{liu2022petpolyhistor, moon2026wimfris}.
Unlike full fine-tuning, which risks distorting pre-trained feature spaces and incurring prohibitive computational costs, PET methods achieve competitive performance with significantly reduced memory and storage overhead.
Current PET literature is generally categorized into three groups: (1) introducing additional lightweight modules with newly learnable parameters \cite{li2021prefix, zhou2022learning}; (2) selectively updating a small subset of parameters within the pre-trained models \cite{guo2021parameter, zaken2022bitfit}; and (3) applying low-rank factorization to the trainable weights, as represented by LoRA \cite{hu2022lora}.
While these techniques have demonstrated success in classification and generation tasks, dense vision-language pixel-level prediction (\textit{e.g.},~RIS) imposes more rigorous demands on cross-modal information fusion and spatial cognition.
Recently, some pioneering works have attempted to adapt PET for RIS, utilizing modular adapters to bridge visual and linguistic modalities.
For instance, ETRIS \cite{xu2023etris} leveraged vision-language bridges to refine intermediate features, while BarLeRIa \cite{wang2024barleria} employed global shortcut tuning to incorporate prior regularization in parallel with the backbone.
Furthermore, DETRIS \cite{huang2025densely} enhanced visual feature propagation through dense interconnections, facilitating cross-modal interaction even within unaligned backbones such as DINO \cite{caron2021dino, oquabdinov2}.
Despite these advancements, effectively adapting PET methods to the remote sensing domain remains a significant challenge due to the substantial domain gap between natural and remote sensing imagery.

Few studies have explored the integration of PET into the RRSIS task.
As a pioneering work, RSRefSeg \cite{chen2025rsrefseg} employed low-rank factorization techniques to adapt the pre-trained SigLIP \cite{zhai2023sigmoidlip} model for visual-textual representation learning in remote sensing imagery, and further leveraged the foundation model SAM \cite{kirillov2023sam} to transform multimodal representations into prompts for mask prediction.
Building upon this framework, RSRefSeg2 \cite{chen2025rsrefseg2} introduced a dual-stage decoupled architecture with implicit cascaded reasoning, enabling coarse target localization followed by refined mask segmentation for prompt generation.
Differently, Aurora \cite{yin2026aurora} introduced a bridger-based design similar to ETRIS, while further enhancing the framework with multi-scale feature fusion following cross-modal alignment.
However, existing methods frequently overlook the hierarchical nature of aerial imagery, where the extreme scale variance and dense visual clutter necessitate specialized intra- and inter-modality processing that goes beyond basic, layer-wise alignment.
Current adapters often fail to effectively aggregate and contextualize features across depths, resulting in a decoupling of spatial precision from linguistic guidance.
To overcome these limitations, our work extends PET to the RRSIS domain by proposing an architecture that prioritizes selective, target-specific feature integration via dual-encoder adaptation, enabling efficient cross-modal alignment while addressing the scale diversity and complex spatial characteristics of remote sensing imagery.


\section{Method}
\label{sec: methodology}
\begin{figure}[t]
  \centering
  \includegraphics[width=0.95\linewidth]{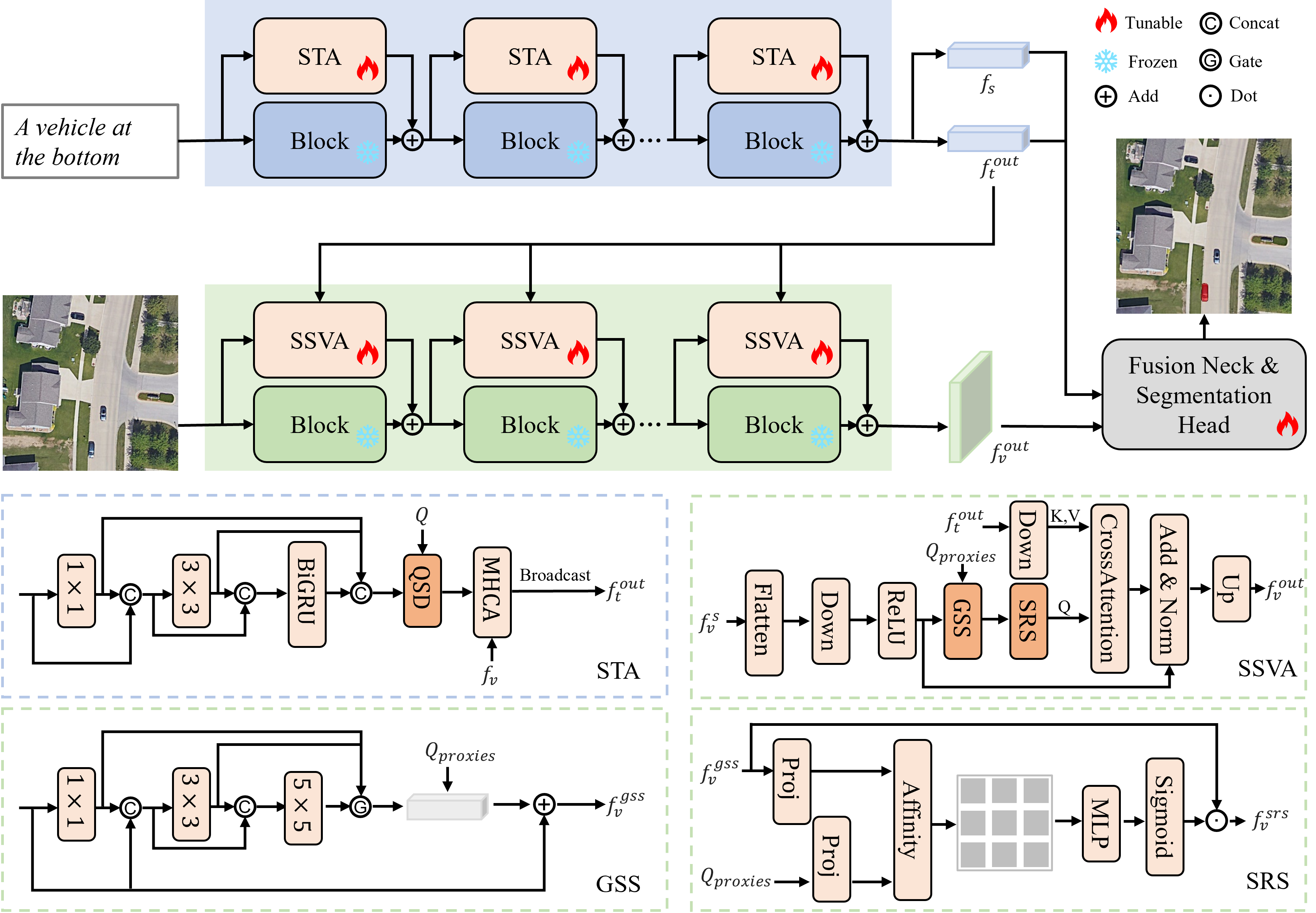}
  \caption{Overview of our proposed S$^4$ECA.
  S$^4$ECA adopts a dual-encoder adaptation paradigm consisting of a semantic-driven textual adapter (STA) and a language-guided scale and spatial selection visual adapter (SSVA).
  STA enhances local and global linguistic dependencies and employs query-based semantic distillation (QSD) to derive compact semantic proxies, which facilitate early grounding through efficient cross-modal interaction.
  SSVA dynamically selects informative receptive fields via gated scale selection (GSS) and suppresses irrelevant regions through spatial reweighting selection (SRS), yielding semantically focused visual representations for subsequent alignment.
  The adapted features are then fused by a simple fusion neck and decoded by a segmentation head to generate the target mask.
  For clarity, their detailed structures are omitted.
  }
  \label{fig: overall framework}
\end{figure}

The overall architecture of our proposed S$^4$ECA framework is illustrated in Figure~\ref{fig: overall framework}.
The proposed framework adopts a dual-encoder adaptation strategy, where dedicated adapters are incorporated into both the visual and textual branches to enable cross-modal interaction and inject task-specific semantic, scale, and spatial priors during feature encoding.
The remainder of this section details the feature extraction process and the proposed adaptation modules, followed by the multimodal fusion neck and the segmentation head for mask prediction.

\subsection{Input Feature Extraction}
To establish a robust foundation for cross-modal alignment, the proposed framework extracts hierarchical representations from both natural language expressions and aerial images.
In alignment with the PET paradigm, we freeze all the parameters of the pre-trained foundation backbones, thereby preserving their original, highly generalizable feature spaces and minimizing computational overhead.

\label{sec: feature extraction}
\noindent
\textbf{Text Encoder}.
Given an input natural language expression $T$, we utilize the pre-trained text transformer from the CLIP \cite{radford2021clip} model to extract rich linguistic features.
Specifically, the text encoder outputs two complementary levels of linguistic embeddings: \textit{word-level features} $f_t \in \mathbb{R}^{L \times C_t}$ and \textit{sentence-level features} $f_s \in \mathbb{R}^{C_t}$.
The $f_t$ embeddings preserve fine-grained token-level semantics across a sequence length of $L$ tokens, where $C_t$ represents the textual channel dimension.
They capture explicit entity types, attributes, and relative spatial cues (\textit{e.g.},~``the red small vehicle to the right'') essential for pixel-level grounding.
Differently, extracted from the final \textit{end-of-text} token, the $f_s$ embedding provides a holistic, global summary of the entire expression, which offers macro-level contextual guidance to constrain the feature search space.

\noindent
\textbf{Image Encoder}.
For the visual branch, we employ DINOv2 \cite{oquabdinov2} based on ViT as the pre-trained vision backbone.
DINOv2 is a self-supervised foundation model trained via discriminative self-distillation, enabling it to capture robust, high-granularity visual semantics and geometric cues.
Given an input remote sensing image $I \in \mathbb{R}^{H \times W \times 3}$, it is first tokenized into non-overlapping patches.
As the input passes through the $N$ transformer layers of the visual backbone, it yields a sequence of multi-level feature maps.
To fully exploit these hierarchical representations for cross-scale reasoning, we extract feature maps from two intermediate layers and the final layer of the backbone, denoted as: $\{f_v^m, f_v^n, f_v^N\}$, where $m, n \in \{1, 2, \dots, N - 1\}$ and $m \neq n$.
Although DINOv2 exhibits strong representation capability and consistently outperforms vanilla CLIP in dense visual perception tasks, its pure self-supervised vision pre-training paradigm does not explicitly establish alignment with the language modality.
Consequently, a notable discrepancy exists between the visual and textual feature spaces when the model is applied to multimodal downstream tasks such as RRSIS.

\subsection{Textual Adapter}
\label{sec: textual adapter}
While the CLIP text encoder provides robust global linguistic representations, its transformer architecture inherently prioritizes long-range dependencies, often overlooking fine-grained local relationships between neighboring tokens.
Furthermore, there exists a significant distribution gap between the CLIP-aligned text embeddings and the self-supervised DINOv2 visual features.
To address these limitations and facilitate early grounding, we introduce the Semantic-driven Textual Adapter (STA), as illustrated in Figure~\ref{fig: overall framework}.

The STA transforms the static textual input into a dynamic, adapted representation through three primary stages: local-global feature enrichment (LGE), query-based semantic distillation (QSD), and cross-modal propagation (CMP).
Specifically, given the word-level features $f_t$, we first enhance the local context using a 1D convolutional block, specifically designed for processing textual data.
This block consists of a $1 \times 1$ convolution to project feature channels, followed by a $3 \times 3$ convolution to broaden the receptive field and capture local neighborhood dependencies.
To capture long-range sequential dependencies bidirectionally, we pass the enriched features through a Bidirectional Gated Recurrent Unit \cite{cho2014bigru}:
\begin{equation}
    \tilde{f}_t = \text{BiGRU}(\text{Conv}_{3\times3}(\text{Conv}_{1\times1}(f_t))),
\end{equation}
This process ensures that each token embedding is conditioned not only on its immediate neighbors but also on the global context of the entire expression.
Inspired by Q-Former \cite{li2023blip2}, we define a set of $Z$ learnable query tokens $Q \in \mathbb{R}^{Z \times C_t}$ to compress the linguistic information into a compact set of task-specific language proxies for semantic distillation.
These queries interact with the enriched features $\tilde{f}_t$ via a cross-attention mechanism to distill the most relevant semantic information:
\begin{equation}
    Q_{proxies} = \text{CrossAtt}(Q, \tilde{f}_t, \tilde{f}_t),
\end{equation}
By constraining the number of queries ($Z \ll L$), we create a \textit{semantic bottleneck}.
This design forces the model to ignore redundant or non-visual words and focus on discriminative language proxies (\textit{e.g.},~red, baseball court, right).
Unlike prior methods (\textit{e.g.},~DETIRS \cite{huang2025densely}) that rely on intra-modal self-attention, our model forces the queries to represent the language semantics in a space optimized for visual alignment.
To achieve early grounding, we perform simple cross-attention between the compressed proxies $Q_{proxies}$ and the corresponding visual features $f_v$:
\begin{equation}
    \hat{Q}_{proxies} = \text{CrossAtt}(Q_{proxies}, f_v, f_v),
\end{equation}
This interaction maps high-level linguistic concepts directly onto spatial visual regions.
Finally, to restore the sequence length, the grounded proxies are propagated back to the original word-level embeddings using a broadcast attention mechanism:
\begin{equation}
    f_t^{out} = \text{Broadcast}(f_t, \hat{Q}_{proxies}),
\end{equation}
Through these progressive operations, the resulting features $f_t^{out}$ serve as a high-fidelity textual proxy, effectively bridging the modality gap by fusing fine-grained token-level information with latent visual-spatial grounding.
This design significantly reduces computational overhead compared to the exhaustive token-to-pixel attention used in previous methods (\textit{e.g.},~Aurora \cite{yin2026aurora}), as the cross-modal interaction is performed only on $Z$ compressed proxies rather than $L$ raw tokens.

\subsection{Visual Adapter}
\label{sec: visual adpater}
In dense mask prediction tasks such as RIS, the efficacy of the visual adapter fundamentally dictates the overall performance of a PET method.
While existing methods emphasize visual adaptation, they predominantly rely on straightforward feature fusion paradigms.
For instance, prior works (\textit{e.g.},~DETRIS\cite{huang2025densely}, MiCA \cite{li2026mica}, and Aurora \cite{yin2026aurora}) attempt to capture rich semantic hierarchies by directly concatenating multi-scale visual features.
However, these multi-scale representations are often highly redundant and do not contribute equally to cross-modal alignment.
Furthermore, driven by the specific linguistic context, only a subset of visual patches is genuinely relevant to the referent.
Indiscriminately aggregating these features introduces both scale redundancy and spatial token redundancy, exacerbating the domain gap between generic visual representations and complex, high-resolution aerial imagery.
To resolve the discrepancies between fine-grained visual cues and semantic context, we propose the language-guided Scale and Spatial Selection Visual Adapter (SSVA).
As illustrated in Figure~\ref{fig: overall framework}, SSVA selectively identifies and refines visual features through three sub-modules: gated scale selection (GSS), spatial reweighting selection (SRS), and language-guided visual adaptation (LVA).

\noindent
\textbf{Gated Scale Selection}.
Specifically, GSS dynamically selects the most informative receptive fields conditioned on the semantic context, rather than statically aggregating features across all scales.
Given an intermediate visual feature map $f_v^s$, where $s \in \{m, n, N\}$, we employ a downsampling operation of linear projection followed by dense convolution branches to capture localized contexts at varying scales:
\begin{equation}
    \begin{array}{l}
    F_v^{s} = \text{ReLU}(\text{Linear}_\text{down}(f_v^s)),\\
    F_{v1}^{s} = \text{Conv}_{1 \times 1}(F_v^s),\\
    F_{v3}^{s} = \text{Conv}_{3 \times 3}([F_v^s, F_{v1}^{s}]),\\
    F_{v5}^{s} = \text{Conv}_{5 \times 5}([F_v^s, F_{v1}^{s}, F_{v3}^{s}]),\\
    \end{array}
\end{equation}
where $[,]$ denotes concatenation along the channel dimension.
The $1 \times 1$ convolution facilitates channel fusion and dimensionality reduction, while the $3 \times 3$ and $5 \times 5$ convolutions extract medium- and large-scale textural details and edge geometries.
As the relevance of each scale depends strictly on the target described in the text, we introduce a dynamic gating mechanism guided by the compressed language proxies obtained from the textual adapter.
The gating weights $\alpha \in \mathbb{R}^{B \times 3}$ control the relative significance of the three dense branches and are formulated as:
\begin{equation}
    \alpha = \text{Softmax}(W_\alpha Q_{proxies} + b_\alpha),
\end{equation}
where $W_\alpha$ and $b_\alpha$ denote the learnable weight matrix and bias vector of a projection layer, respectively.
The gated multi-scale features are then fused and added to the original features via a residual connection:
\begin{equation}
    F_v^{gss} = F_v^s + \alpha_1 \cdot F_{v1}^{s} + \alpha_2 \cdot F_{v3}^{s} + \alpha_3 \cdot F_{v5}^{s},
\end{equation}
This design paradigm shifts from traditional scale aggregation to \textit{language-guided scale selection}, enabling the model to dynamically calibrate its receptive field based on semantic demands.

\noindent
\textbf{Spatial Reweighting Selection}.
To address spatial information redundancy, SRS evaluates the textual relevance of every visual patch.
We first compute a lightweight cross-modal affinity matrix $S$ between the scale-selected visual features and the language proxies:
\begin{equation}
    S = F_v^{gss} W_v (Q_{proxies} W_t)^T,
\end{equation}
where $W_v$ and $W_t$ are linear projection matrices and $T$ denotes the matrix transpose.
We specifically utilize the compressed language proxies because their condensed, high-dimensional semantic nature makes them superior to both granular word-level and overly broad sentence-level embeddings for affinity modeling.
The affinity matrix is then passed through a two-layer MLP to determine the alignment intensity, followed by a sigmoid normalization function $\sigma$.
The spatial weights are subsequently applied element-wise to the visual features:
\begin{equation}
    F_v^{srs} = F_v^{gss} \odot (\sigma(\text{MLP}(S))),
\end{equation}
Note that SRS reweights each visual token according to its relevance to the textual query while preserving the full spatial grid. This design naturally maintains the spatial geometry encoded in the original feature map without introducing explicit coordinate embeddings, allowing the model to suppress background clutter and emphasize target-relevant regions without compromising spatial precision.

\noindent
\textbf{Language-guided Visual Adaptation}.
Following scale and spatial selection, LVA executes a deep cross-modal alignment using a cross-attention mechanism.
Here, the refined visual features act as queries, while the dense word-level features are injected into the visual embedding space as keys and values.
This allows every visual feature point to establish relationships with fine-grained semantic tokens from the text.
To mitigate potential information loss during this dynamic reconstruction, the attention output is integrated using residual connections, dropout, and Layer Normalization.
\begin{equation}
    f_{v}^{lva} = \text{CrossAtt}(F_v^{srs}, f_t^{out}, f_t^{out}) + F_v^{s},
\end{equation}
Finally, an upsampling projection layer restores the reconstructed visual features to the target output dimension:
\begin{equation}
    f_v^{out} = \text{Linear}_\text{up}(f_{v}^{lva}),
\end{equation}
Through the consecutive application of GSS, SRS, and LVA, our proposed SSVA successfully executes cross-modal visual adaptation.
By mitigating both scale and spatial redundancy through targeted dynamic selection mechanisms, it guarantees that the visual representations are both highly discriminative and strictly aligned with the referring expression.

\subsection{Fusion Neck}
\label{sec: neck}
To effectively aggregate the adapted visual and linguistic representations, we employ a hierarchical, two-stage fusion neck on the basis of ETRIS \cite{xu2023etris}.
This architecture facilitates a progressive transition from coarse global localization to fine-grained pixel-level segmentation.
The neck is divided into two sequential operations: global language-guided prior localization (GLPL) and bidirectional cross-modal deep perception (BCDP).

The primary objective of this stage is to narrow down the visual search space by injecting macro-level semantic constraints into the multi-scale visual features.
Given the adapted multi-scale visual features $f_v^{out}$ and the global sentence-level feature $f_s$, we first map them to a shared embedding space.
We then apply a cross-attention mechanism, where the visual features act as queries and the sentence embedding serves as keys and values:
\begin{equation}
    F_{p} = \text{CrossAtt}(f_v^{out}, f_s, f_s),
\end{equation}
This operation allows the global text prompt to act as a spatial prior, anchoring the visual features around potentially relevant target regions.
To resolve the intense spatial variations typical of complex remote sensing scenes, we supplement $F_p$ with explicit positional awareness.
We pass the initial fused features through a coordinate convolution layer \cite{liu2018coord}, which appends explicit pixel coordinate channels to model long-range spatial spatial dependencies.
Finally, to consolidate the coordinate-aware features and eliminate local noise, the cross-attention output and the enhanced coordinate features are concatenated along the channel dimension and projected using a standard $3 \times 3$ convolutional layer, yielding the coarse position-aware visual features:
\begin{equation}
    \begin{array}{l}
    F_{coord} = \text{CoordConv}(F_{p}) + F_{p},\\
    F_{c} = \text{Conv}_{3\times3}([F_p, F_{coord}]),
    \end{array}
\end{equation}

Once coarse localization is established, this stage models the complex, dense dependencies between individual visual tokens and fine-grained word embeddings to ensure precise delineation.
The step is accomplished via a dual-path bidirectional paradigm.
The position-aware visual features $F_c$ is flattened spatially into a sequence of tokens and concatenated with the adapted word-level embeddings $f^{out}_t$.
This heterogeneous sequence is passed through a self-attention layer to jointly model token and patch relationships.
To further represent text-conditioned visual semantics, the self-attended output is treated as a query matrix and cross-aligned back against the adapted word embeddings to obtain the fused visual features $F^{fuse}_v$.
In parallel, to prevent the textual features from drifting during optimization, the global sentence feature $f_s$ is used as a query to probe $F^{fuse}_v$ for the fused textual features $F^{fuse}_t$.

\subsection{Segmentation Head}
\label{sec: head}
To ensure a fair comparison, we follow ETRIS \cite{xu2023etris}, incorporating a mainstream referring image segmentation head.
Specifically, the visually-aware linguistic features $F^{fuse}_t$ are projected through a linear layer to generate a target-specific weight matrix $W_h$ and a bias vector $b_h$:
\begin{equation}
    W_h, b_h = \text{Linear}(F^{fuse}_t),
\end{equation}
where $W_h$ represents the dynamic convolutional kernel and $b_h$ denotes its corresponding bias term.
To preserve and recover precise boundaries, we first apply a bilinear upsampling operation to restore spatial fidelity.
Following the resolution recovery, the text-guided dynamic convolution operation is executed pixel-wise across the enhanced features to project the multi-modal features into a dense logit space:
\begin{equation}
    \hat{M} = \text{Conv2D}(\text{Linear}_\text{up}(F^{fuse}_v), W) + b,
\end{equation}

We employ a hybrid loss function that combines pixel-level classification accuracy with global structural alignment.
The total training objective is formulated as a joint optimization of Binary Cross-Entropy (BCE) loss and Dice loss.
The BCE loss enforces strict pixel-wise supervision, penalizing classification errors independently at each spatial location, while the Dice loss optimizes the structural overlap between the prediction and the target mask:
\begin{equation}
    L_{\text{BCE}} = -\frac{1}{N} \sum_{i=1}^{N} \left[ M_i \log(\hat{M}_i) + (1 - M_i) \log(1 - \hat{M}_i) \right],
\end{equation}
\begin{equation}
    L_{\text{Dice}} = 1 - \frac{2 \sum_{i=1}^{N} M_i \hat{M}_i}{\sum_{i=1}^{N} M_i + \sum_{i=1}^{N} \hat{M}_i},
\end{equation}
where $N$ denotes the total number of pixels in the image, $M_i$ represents the ground-truth label for the $i$-th pixel.
The overall loss function $L$ is defined as a weighted linear combination of both objectives:
\begin{equation}
    L = \lambda_1 L_{\text{BCE}} + \lambda_2 L_{\text{Dice}},
\end{equation}
where $\lambda_1$ and $\lambda_2$ are hyper-parameters tasked with balancing the scales of the two loss terms during the end-to-end optimization process.


\section{Experiments}

\subsection{Experiment Setup}
\label{sec: setup}

\subsubsection{Datasets}
\label{sec: datasets}
To evaluate the effectiveness of our proposed framework, we conducted experiments on two publicly available RRSIS benchmarks: RRSIS-D \cite{liu2024rmsin} and RefSegRS \cite{yuan2024rrsis}.
\begin{itemize}[leftmargin=*, itemsep=0pt]
  \item \textbf{RRSIS-D.}
  This dataset is a large-scale collection derived from the RSVGD \cite{zhan2023rsvg} dataset.
  It consists of 17,402 image-expression-mask triplets, partitioned into training (12,181), validation (1,740), and test (3,481) sets.
  The samples feature 20 distinct semantic categories such as aircraft, golf courses, and stadiums, complemented by seven specific attributes that enhance its semantic complexity.
  The dataset is characterized by significant scale variation, with object sizes ranging from minimal footprints to over 400,000 pixels, and images are resized to 800 $\times$ 800 pixels with spatial resolutions varying between 0.5 and 30 m/pixel.
  \item \textbf{RefSegRS.}
  This dataset incorporates pixel-level annotations sourced from the SkyScapes \cite{azimi2019skyscapes} dataset.
  It provides 4,420 image-expression-mask triplets, divided into a training set (2,172), a validation set (431), and a test set (1,817).
  The dataset covers 14 hierarchical semantic categories, such as buildings, roads, and vehicles, each described by five distinct attributes.
  Unlike RRSIS-D, all images in RefSegRS are captured at a uniform resolution of 0.13 m/pixel and are resized to 512 $\times$ 512 pixels, providing a consistent high-resolution environment for evaluating referring capabilities of models.
\end{itemize}

\subsubsection{Evaluation Protocol}
In accordance with established protocols in the field \cite{liu2024rmsin, yang2022lavt, li2025sbanet}, we utilized three primary metrics to quantitatively assess segmentation performance: mean Intersection over Union (mIoU), overall Intersection over Union (oIoU), and Precision at thresholds $X$ (Pr@$X$, where $X \in \{0.5, 0.7, 0.9\}$).
mIoU calculates the average IoU across all individual test samples, providing an assessment that treats large and small objects with equal weight.
oIoU is computed as the ratio of the total intersection area to the total union area across the entire test set.
By aggregating the areas, this metric inherently places greater emphasis on larger objects.
Pr@$X$ serves as a sample-level evaluation metric, measuring the percentage of samples that achieve an IoU score exceeding a specified threshold $X$.

\subsubsection{Implementation Details}
\label{sec: details}
We implemented our framework using the PyTorch \cite{paszke2019pytorch} library.
To ensure robust feature representation, we utilized the CLIP \cite{radford2021clip} text encoder for textual processing, while the DINOv2 \cite{oquabdinov2} visual backbone was employed for image encoding.
Notably, we adopted an end-to-end optimization strategy.
Following prior studies \cite{xu2023etris, huang2025densely}, all input images were resized to 448 $\times$ 448 pixels, and training was performed using the Adam \cite{kingma2014adam} optimizer with a total duration of 65 epochs.
To stabilize convergence, the learning rate underwent a decay at epoch 35.
We evaluated two model variants, S$^4$ECA-B and S$^4$ECA-L, to assess the scalability of our approach.
S$^4$ECA-B utilized the DINOv2-B/14 backbone, comprising a 12-layer Transformer \cite{vaswani2017attention} with our proposed SSVA adapters integrated into layers $[2, 4, 6, 8, 10, 12]$.
This variant was trained on two NVIDIA L40S GPUs with an initial learning rate of $0.0005$, which decayed to $0.0001$.
Similarly, S$^4$ECA-L employed the DINOv2-L/14 backbone, a 24-layer Transformer architecture, with SSVA adapters positioned at layers $[4, 8, 12, 16, 20, 24]$.
This model was trained on two NVIDIA H100 GPUs, using an initial learning rate of $0.0003$ that also decayed to $0.0001$.
The specific placement of these adapters was determined through empirical validation, balancing segmentation performance with computational overhead.
We set the weighting hyperparameters to $\lambda_1 = 1$ and $\lambda_2 = 1$ to maintain an equal balance between the two loss components.



\subsection{Comparison with the State-of-the-Art}
\label{sec: comparison}

\begin{table*}\scriptsize
  \caption{Comparison with existing RIS methods on the RRSIS-D \cite{liu2024rmsin} dataset in terms of oIoU, mIoU, and Pr@$k$.
  The short names for different visual encoders (\textit{e.g.},~ResNet \cite{he2016deepresnet}, Swin Transformer\cite{liu2021swin}, and SigLIP \cite{zhai2023sigmoidlip} plus SAM-L \cite{kirillov2023sam}) are defined as: R-101, Swin-B and SigLIP+L.
  The best results are \textbf{bold}.
  Note that the performance results of previous methods are taken directly from either SBANet \cite{li2025sbanet} or their original works, when available; otherwise, the missing results are marked with $\dagger$ to indicate the corresponding re-implementations.
  }
  \centering
  \setlength\tabcolsep{3pt}
  \begin{tabular}{p{3.2cm}cccccccccccc}
  \toprule
  \multirow{2}*{Method} & \multicolumn{1}{c}{Vis} & \multicolumn{1}{c}{Tex} & \multicolumn{2}{c}{Pr@0.5} & \multicolumn{2}{c}{Pr@0.7} & \multicolumn{2}{c}{Pr@0.9} & \multicolumn{2}{c}{oIoU} & \multicolumn{2}{c}{mIoU}\\
 & Enc. & Enc. & Val & Test & Val & Test & Val & Test & Val & Test & Val & Test\\
 \midrule
 \multicolumn{13}{l}{\textit{Traditional Full Fine-tuning}} \\
 \midrule
  RRN\cite{li2018rrn} \tiny{\textit{cvpr18}}   & R-101 & LSTM & 51.09 & 51.07 & 33.04 & 32.77 & 6.14 & 6.37 & 66.53 & 66.43 & 46.06 & 45.64\\
  CMSA\cite{ye2019csma} \tiny{\textit{cvpr19}}  & R-101 & None & 55.68 & 55.32 & 38.27 & 37.43 & 9.02 & 8.15 & 69.68 & 69.39 & 48.85 & 48.54\\
  CMPC\cite{huang2020cmpc} \tiny{\textit{cvpr20}} & R-101 & LSTM & 57.93 & 55.83 & 38.50 & 36.94 & 9.31 & 9.19 & 70.15 & 69.22 & 50.41 & 49.24\\
  CMPC+ \cite{liu2021cmpc+} \tiny{\textit{tpami21}} & R-101 & LSTM & 59.19 & 57.65 & 49.36 & 36.97 & 8.16 & 7.78 & 70.14 & 68.64 & 51.41 & 50.24\\
  CRIS \cite{wang2022cris} \tiny{\textit{cvpr22}} & R-101 & CLIP & 56.44 & 54.84 & 39.77 & 38.06 & 11.84 & 11.52 & 70.98 & 70.46 & 50.75 & 49.69\\
  LAVT \cite{yang2022lavt} \tiny{\textit{cvpr22}} & Swin-B & BERT & 69.54 & 69.52 & 53.16 & 53.29 & 24.25 & 24.94 & 77.59 & 77.19 & 61.46 & 61.04\\
  CARIS \cite{liu2023caris} \tiny{\textit{mm23}} & Swin-B & BERT & 71.61 & 71.50 & 54.14 & 52.92 & 23.79 & 23.90 & 77.48 & 77.17 & 62.88 & 62.12\\
  LGCE \cite{yuan2024rrsis} \tiny{\textit{tgrs24}} & Swin-B & BERT & 68.10 & 67.65 & 
  52.24 & 51.45 & 23.85 & 23.33 & 76.68 & 76.34 & 60.16 & 59.37\\
  RMSIN \cite{liu2024rmsin} \tiny{\textit{cvpr24}}  & Swin-B & BERT & 74.66 & 74.26 & 57.41 & 55.93 & 24.43 & 24.53 & 78.27 & 77.79 & 65.10 & 64.20\\
  DANet \cite{pan2024mmrrsis} \tiny{\textit{mm24}}  & Swin-B & BERT & 73.69 & 74.31 & 57.92 & 56.13 & 27.05 & 24.11 & 79.85 & 77.86 & 66.07 & 64.33\\
  FIANet \cite{lei2024exploringarxiv} \tiny{\textit{tgrs25}} & Swin-B & BERT & 74.32 & 74.46 & 57.22 & 56.31 & 25.10 & 24.13 & 78.43 & 76.91 & 65.02 & 64.01\\
  LSCF \cite{ma2025lscf} \tiny{\textit{tgrs25}}  & Swin-B & BERT & 75.17 & 74.30 & 57.99 & 56.32 & 25.98 & 25.67 & 78.14 & 77.42 & 65.15 & 64.25\\
  SBANet \cite{li2025sbanet} \tiny{\textit{isprs p\&rs25}} & Swin-B & BERT  & 76.84 & 75.91 & 58.86 & 57.05 & 26.70 & 25.38 & 80.02 & \textbf{79.22} & 66.71 & 65.52\\
  HTVG \cite{zhang2026htvg}  \tiny{\textit{pr26}} & Swin-B & BERT & 75.69 & 75.21 & 57.87 & 57.45 & 26.21 & 24.96 & 78.88 & 77.85 & 65.43 & 64.82\\
  CroBIM-U \cite{sun2026crobim-u} \tiny{\textit{tgrs26}} & ConvNeXt-B & BERT & 76.31 & 75.60 & 57.67 & 56.47 & 24.89 & 24.16 & 77.83 & 76.70 & 66.07 & 65.07\\
  RS2-SAM2 \cite{rong2026rs2sam2} \tiny{\textit{aaai26}} & SAM2-L & BEiT-3 & \textbf{79.25} & \textbf{77.56} & \textbf{63.85} & \textbf{61.76} & \textbf{30.40} & \textbf{29.73} & \textbf{80.16} & 78.99 & \textbf{68.81} & \textbf{66.72}\\

  \midrule
  \multicolumn{13}{l}{\textit{Parameter-Efficient Tuning Low-rank-based}} \\
  \midrule
  RSRefSeg \cite{chen2025rsrefseg} \tiny{\textit{igarss25}} & SigLIP+L & SigLIP & - & 74.49 & - & 58.73 & - & 30.80 & - & 77.24 & - & 64.67\\
  RSRefSeg2 \cite{chen2025rsrefseg2} \tiny{\textit{tgrs26}} & SigLIP2+L & SigLIP2 & - & 80.23 & - & 65.41 & - & 31.05 & - & 79.45 & - & 69.17\\
  
  \midrule
  \multicolumn{13}{l}{\textit{Parameter-Efficient Tuning Adapter-based}} \\
  \midrule

  ETRIS \cite{xu2023etris}  \tiny{\textit{iccv23}$^\dagger$} & CLIP ViT-B & CLIP & 61.55 & 60.47 & 43.07 & 40.38 & 12.38 & 11.01 & 72.13 & 70.87 & 54.63 & 53.50\\
  BarLeRIa \cite{wang2024barleria} \tiny{\textit{iclr24}$^\dagger$} & CLIP ViT-B & CLIP & 66.91 & 64.26 & 48.19 & 47.81 & 18.40 & 17.71 & 74.69 & 73.31 & 58.65 & 57.81\\
  DETRIS \cite{huang2025densely} \tiny{\textit{aaai25}$^\dagger$} & DINOv2-B & CLIP & 72.98 & 71.77 & 55.45 & 53.36 & 24.41 & 23.80 & 76.78 & 76.27 & 63.41 & 62.81\\
  Aurora \cite{yin2026aurora} \tiny{\textit{icaspp26}} & Swin-B & BERT  & 75.91 & 75.80 & 58.44 & 57.42 & 26.77 & 25.90 & 79.25 & 78.72 & 66.19 & 65.90\\
  S$^4$ECA-B (Ours) & DINOv2-B & CLIP & 79.16 & 78.28 & 62.73 & 61.10 & 30.74 & 30.03 & 79.63 & 78.75 & 68.92 & 68.43\\
  S$^4$ECA-L (Ours) & DINOv2-L & CLIP & \textbf{81.45} & \textbf{81.03} & \textbf{64.54} & \textbf{63.12} & \textbf{32.14} & \textbf{31.73} & \textbf{81.19} & \textbf{80.07} & \textbf{70.29} & \textbf{69.76}\\
  \bottomrule
  \end{tabular}
  \vspace{-2mm}
  \label{tab:comparison with SOTA}
\end{table*}

\subsubsection{Quantitative Results on the RRSIS-D Dataset}
\label{sec: results on rrsis}
The quantitative evaluation results on the RRSIS-D dataset are summarized in Table~\ref{tab:comparison with SOTA}.
Overall, our proposed S$^4$ECA framework achieved state-of-the-art (SOTA) performance across the various metrics, substantially outperforming existing full fine-tuning methods and PET approaches.
Among the traditional full fine-tuning methods, early approaches such as RRN \cite{li2018rrn} and CMPC \cite{huang2020cmpc} exhibited limited performance due to their relatively weak visual representations and shallow cross-modal interactions.
Benefiting from stronger vision-language architectures, recent methods consistently improved segmentation accuracy.
In particular, RS2-SAM2 \cite{rong2026rs2sam2} achieved the strongest performance among full fine-tuning methods, owing to its large-scale SAM2-L \cite{ravi2025sam2} backbone and prompt-based mask generation strategy.
Nevertheless, our base model, S$^4$ECA-B, yielded a test mIoU of 68.43\%, which surpasses existing fully fine-tuned remote sensing models like SBANet \cite{li2025sbanet} (65.52\%) and CroBIM-U \cite{sun2026crobim-u} (65.07\%).
When scaling up to a large backbone, S$^4$ECA-L established a new performance ceiling.
For instance, it surpassed the previous heavily parameterized SOTA model, RS2-SAM2, by 1.08\% in test oIoU and by 3.04\% in test mIoU.
These results demonstrate that effective parameter-efficient adaptation can fully exploit the transferable knowledge of foundation models without requiring full-parameter optimization.
Within the adapter-based PET category, S$^4$ECA consistently outperformed all competing methods by a substantial margin.
Compared with Aurora \cite{yin2026aurora}, the previous best-performing method, S$^4$ECA-L improved the test Pr@0.5 and mIoU by 5.23\% and 3.86\%, respectively.
These gains verify the effectiveness of the proposed dual-encoder adaptation framework.
While Aurora and DETRIS \cite{huang2025densely} blindly aggregate multi-scale features, S$^4$ECA dynamically filters out scale and spatial redundancies, ensuring the network processes only target-relevant features.
As a result, the proposed S$^4$ECA establishes stronger cross-modal correspondence and produces more discriminative multimodal representations for RRSIS.
We also evaluated our model against recent low-rank-based PET frameworks for RRSIS, namely RSRefSeg \cite{chen2025rsrefseg} and RSRefSeg2 \cite{chen2025rsrefseg2}.
Although RSRefSeg2 relies on massive, multiple external foundation backbones (\textit{e.g.},~SigLIP \cite{zhai2023sigmoidlip} and SAM \cite{kirillov2023sam}) combined with a highly complex, two-stage pipeline, S$^4$ECA-L still achieved higher test oIoU and mIoU scores.
These results demonstrate that explicitly mitigating scale, spatial redundancies via our selection mechanism within a unified end-to-end framework is vastly more effective for cross-modal mask predictions than stacking external heavy vision-language encoders.
Overall, the superior results validate the effectiveness of jointly modeling semantic, scale, and spatial priors for parameter-efficient cross-modal adaptation in remote sensing imagery.

\begin{table}\scriptsize
  \caption{Comparison with existing RIS methods on the RefSegRS \cite{yuan2024rrsis} test set in terms of oIoU, mIoU, and Pr@$k$.
  The best results are \textbf{bold}.
  }
  \centering
  \setlength\tabcolsep{3pt}
  \begin{tabular}{p{3.8cm}ccccc}
  \toprule
  Method & Pr@0.5 & Pr@0.7 & Pr@0.9 & oIoU & mIoU\\
 \midrule
 \multicolumn{6}{l}{\textit{Traditional Full Fine-tuning}} \\
 \midrule
  RRN \cite{li2018rrn} \tiny{\textit{cvpr18}} & 31.21 & 15.30 & 1.10 & 66.12 & 43.34\\
  CMSA \cite{ye2019csma} \tiny{\textit{cvpr19}} & 28.07 & 12.71 & 0.83 & 64.53 & 41.47 \\
  CMPC \cite{huang2020cmpc} \tiny{\textit{cvpr20}} & 26.57 & 11.26 & 0.88 & 61.25 & 33.57 \\
  CMPC+ \cite{liu2021cmpc+} \tiny{\textit{tpami21}} & 51.27 & 29.54 & 3.27 & 68.23 & 54.21 \\
  LAVT \cite{yang2022lavt} \tiny{\textit{cvpr22}} & 71.44 & 32.14 & 4.51 & 76.46 & 57.74\\
  CARIS \cite{liu2023caris} \tiny{\textit{mm23}} & 71.82 & 33.67 & 5.03 & 76.62 & 58.30\\
  LGCE \cite{yuan2024rrsis} \tiny{\textit{tgrs24}} & 73.75 & 39.46 &  5.45 & 76.81 & 59.96\\
  RMSIN \cite{liu2024rmsin} \tiny{\textit{cvpr24}} & 72.26 & 39.37 & 5.38 & 76.29 & 59.63 \\
  DANet \cite{pan2024mmrrsis} \tiny{\textit{mm24}} & 76.61 & 42.72 & 8.04 & 79.53 & 62.14 \\
  FIANet \cite{lei2024exploringarxiv} \tiny{\textit{tgrs25}} &
  76.73 & 42.85 & 8.41 & 78.04 & 61.77 \\
  SBANet \cite{li2025sbanet} \tiny{\textit{isprs p\&rs25}} & 77.02 & 44.15 & 8.97 & 79.86 & 62.73\\
  HTVG \cite{zhang2026htvg} \tiny{\textit{pr26}} & 83.60 &  63.02 &  7.26  & 80.12 &  68.81\\
  CroBIM-U \cite{sun2026crobim-u} \tiny{\textit{tgrs26}} & 76.68 &  34.81 &  3.26  & 73.81 &  60.08\\
  RS2-SAM2 \cite{rong2026rs2sam2} \tiny{\textit{aaai26}} & \textbf{84.31} &  \textbf{70.89} &  \textbf{21.19}  & \textbf{80.87} &  \textbf{73.90}\\

  \midrule
  \multicolumn{6}{l}{\textit{Parameter-Efficient Tuning Low-rank-based}} \\
  \midrule
  RSRefSeg \cite{chen2025rsrefseg} \tiny{\textit{igarss25}$^\dagger$} & 81.14 &  65.77 &  16.06  & 79.69 &  68.71 \\
  RSRefSeg2 \cite{chen2025rsrefseg2} \tiny{\textit{tgrs26}} & 88.22 &  73.97 &  34.40  & 81.24 &  77.39\\

  \midrule
  \multicolumn{6}{l}{\textit{Parameter-Efficient Tuning Adapter-based}} \\
  \midrule
  ETRIS \cite{xu2023etris} \tiny{\textit{iccv23}$^\dagger$} & 45.11 &  24.22 &  1.02  & 68.71 &  45.87\\
  BarLeRIa \cite{wang2024barleria} \tiny{\textit{iclr24}$^\dagger$} & 62.50 & 30.41 & 3.71 & 70.45 & 55.02 \\
  DETRIS \cite{huang2025densely} \tiny{\textit{aaai25}$^\dagger$} & 79.79 & 61.89 & 17.34 & 77.31 & 68.14\\
 
  Aurora \cite{yin2026aurora} \tiny{\textit{icaspp26}} & 59.96 &  27.50 &  2.31  & 72.25 &  53.59\\
  S$^4$ECA-B (Ours) & 85.27 & 69.83 & 20.79 & 80.59 & 75.10\\
  S$^4$ECA-L (Ours) & \textbf{88.51} & \textbf{72.90} & \textbf{25.12} & \textbf{82.35} & \textbf{77.64}\\
  \bottomrule
  \end{tabular}
  \label{tab:comparison with SOTA on refsegrs}
\end{table}

\subsubsection{Quantitative Results on the RefSegRS Dataset}
\label{sec: results on refsegrs}
The quantitative results on the RefSegRS benchmark is reported in Table~\ref{tab:comparison with SOTA on refsegrs}.
Consistent with the observations on RRSIS-D, the proposed S$^4$ECA achieved highly competitive performance across all evaluation metrics and established a new SOAT among adapter-based PET methods.
Compared with traditional full fine-tuning approaches, S$^4$ECA-L achieved the best overall performance in terms of both precision and IoU scores.
Specifically, it surpassed the strongest full fine-tuning baseline, RS2-SAM2, by 1.48\% in oIoU and 3.74\% in mIoU.
Direct comparison within the adapter-based category highlights the precise value of our architectural modifications.
Compared with DETRIS \cite{huang2025densely}, which adopts DINOv2 and CLIP backbones similar to our framework, S$^4$ECA-L improved Pr@0.5 from 79.79\% to 88.51\% and mIoU from 68.14\% to 77.64\%.
Furthermore, S$^4$ECA-B already surpassed all existing adapter-based baselines, highlighting the effectiveness of the proposed textual adaptation and visual selection mechanisms.
Compared to low-rank tuning alternatives, S$^4$ECA-L maintained a clear advantage in overall segmentation quality, outperforming the RSRefSeg2 framework in both oIoU and mIoU by 1.11\% and 0.25\%, respectively.
While our model yielded highly competitive results on the Pr@0.5 and Pr@0.7 metrics, a noticeable performance gap emerged on the more stringent Pr@0.9 threshold.
We attribute this gap primarily to the characteristics of RefSegRS, where a single referring expression may correspond to multiple disconnected objects distributed across distant image regions, as discussed in \cite{li2025sbanet}.
While S$^4$ECA focuses on enhancing global semantic grounding and target-aware feature selection rather than explicit region-wise refinement, small localization errors accumulated across individual instances can substantially reduce the Pr@0.9 score, despite maintaining strong overall segmentation quality as reflected by the superior oIoU and mIoU results.
Incorporating explicit region-level reasoning or instance-aware refinement strategies \cite{liu2023gres} may further improve the segmentation accuracy of disconnected targets.
Nevertheless, the overall performance gains on RefSegRS demonstrates the robustness and generalization capability of the proposed semantic-driven cross-modal adaptation strategy across different remote sensing benchmarks.

\subsubsection{Qualitative Results}
\label{sec: vis}
\begin{figure*}[tp]
  \centering
  \begin{subfigure}[t]{1\linewidth}
  \subcaption*{\textbf{$Expression$:} A chimney on the right}
  \end{subfigure}
 
  \begin{subfigure}[t]{0.155\linewidth}
  \includegraphics[width=1\linewidth]{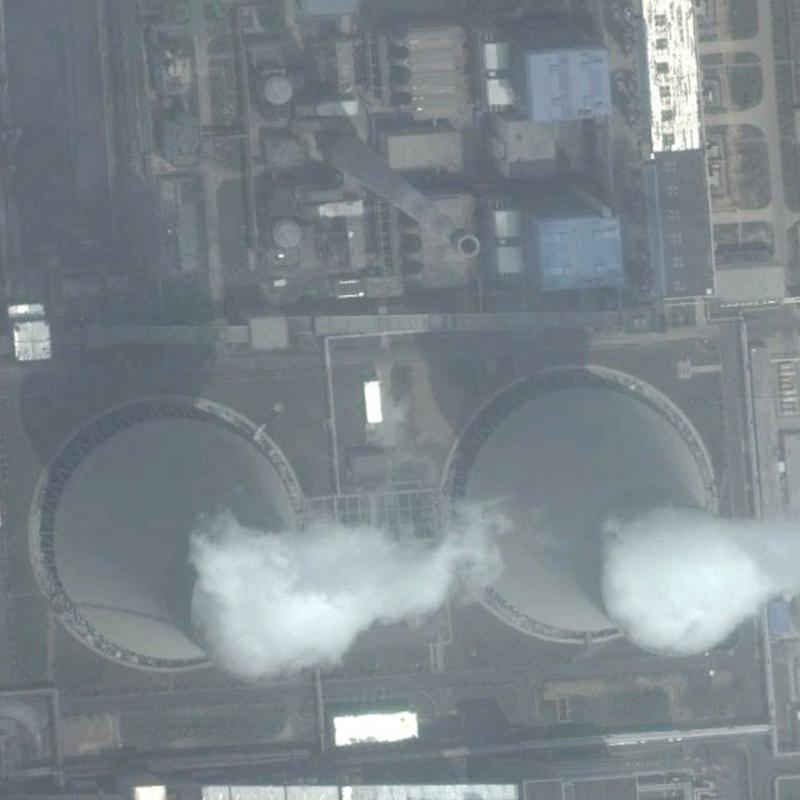}
  \end{subfigure}
  \begin{subfigure}[t]{0.155\linewidth}
  \includegraphics[width=1\linewidth]{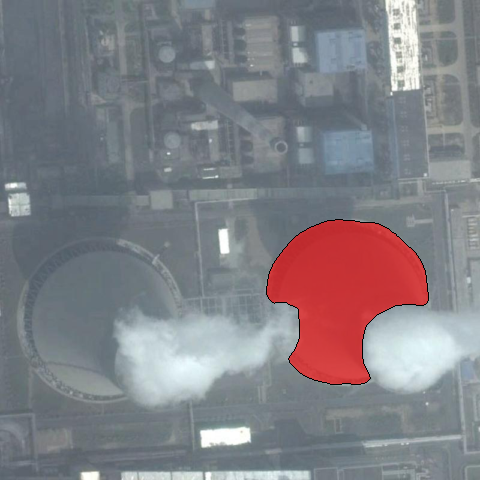}
  \end{subfigure}
  \begin{subfigure}[t]{0.155\linewidth}
  \includegraphics[width=1\linewidth]{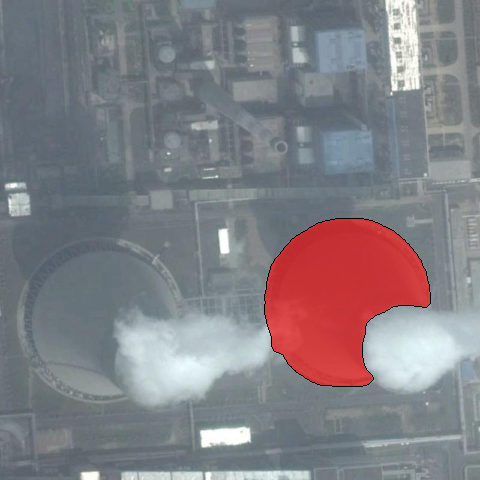}
  \end{subfigure}
  \begin{subfigure}[t]{0.155\linewidth}
  \includegraphics[width=1\linewidth]{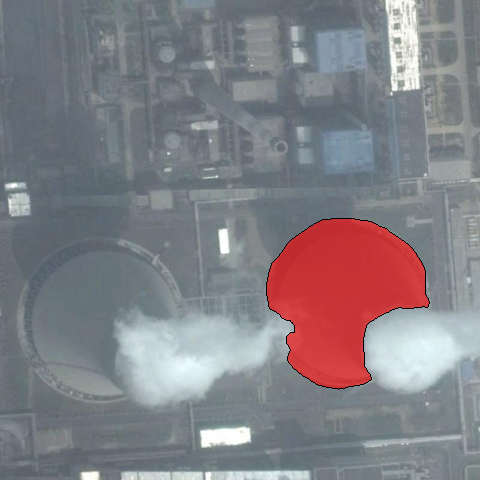}
  \end{subfigure}
  \begin{subfigure}[t]{0.155\linewidth}
  \includegraphics[width=1\linewidth]{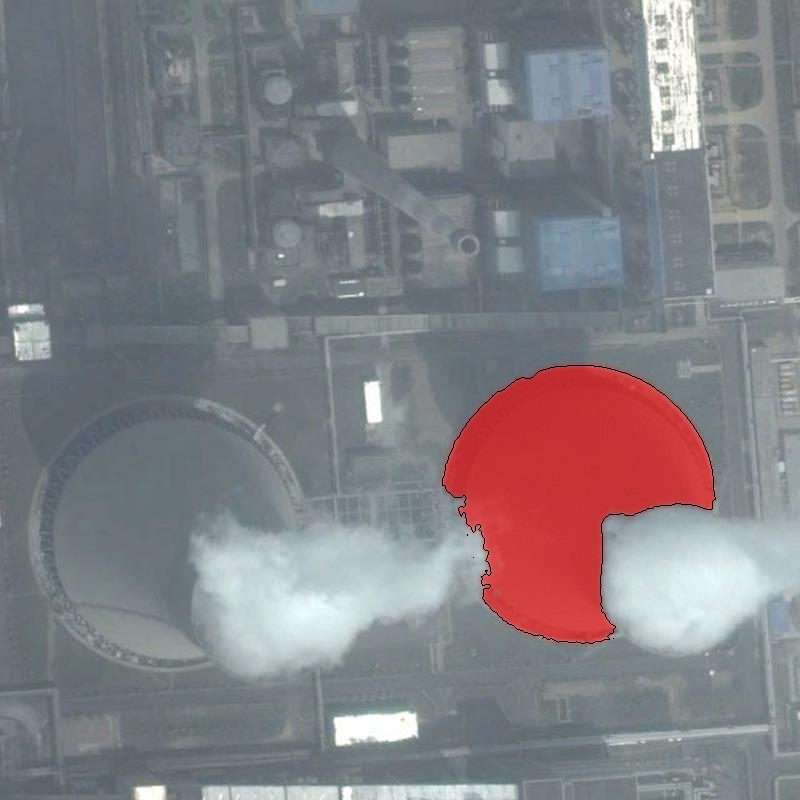}
  \end{subfigure}
  \begin{subfigure}[t]{0.155\linewidth}
  \includegraphics[width=1\linewidth]{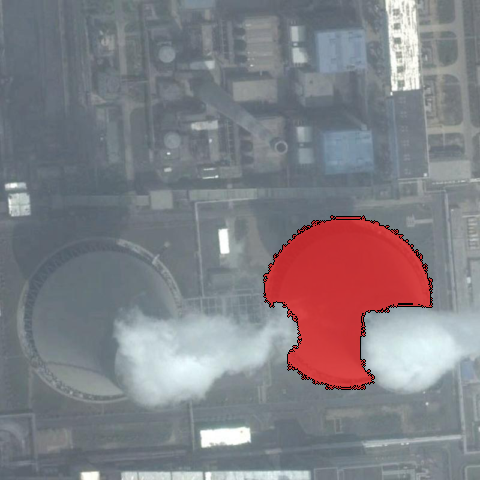}
  \end{subfigure}

  \vspace{-5mm} 

  \begin{subfigure}[t]{1\linewidth}
  \subcaption*{\textbf{$Expression$:} The ship on the far right}
  \end{subfigure}
  
  \begin{subfigure}[t]{0.155\linewidth}
  \includegraphics[width=1\linewidth]{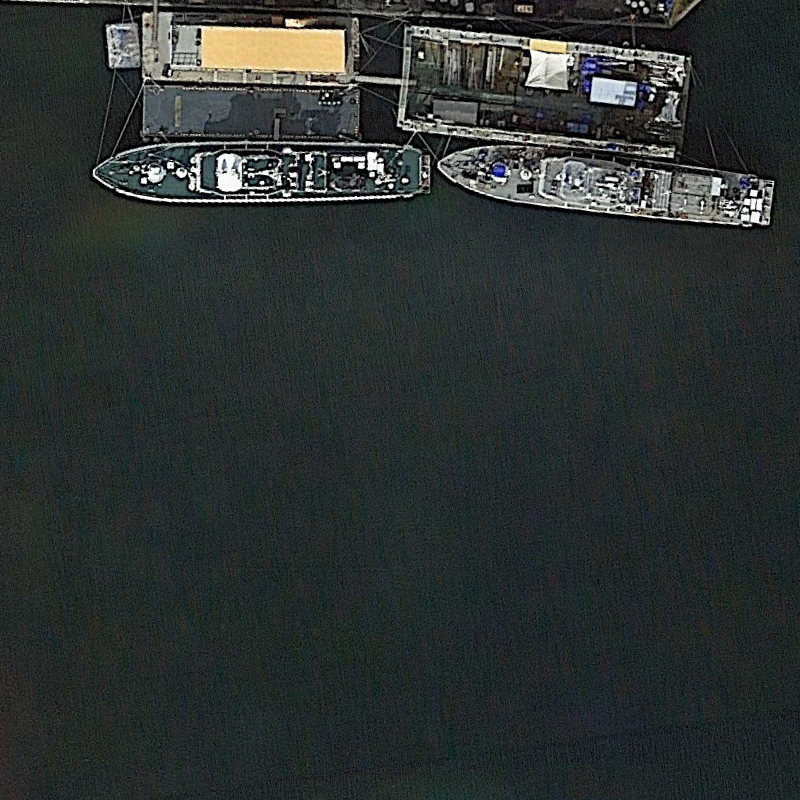}
  \end{subfigure}
  \begin{subfigure}[t]{0.155\linewidth}
  \includegraphics[width=1\linewidth]{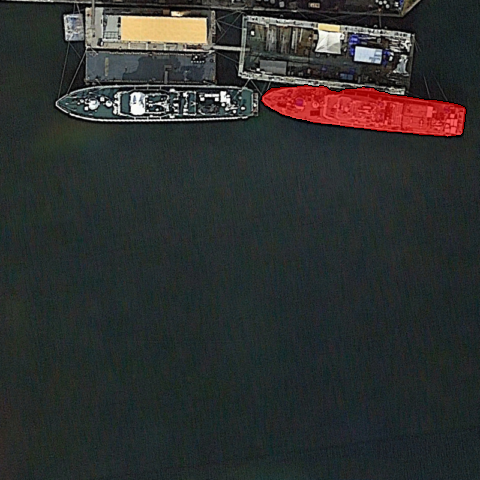}
  \end{subfigure}
  \begin{subfigure}[t]{0.155\linewidth}
  \includegraphics[width=1\linewidth]{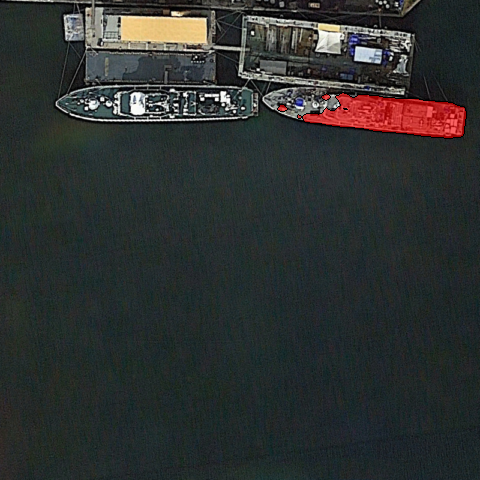}
  \end{subfigure}
  \begin{subfigure}[t]{0.155\linewidth}
  \includegraphics[width=1\linewidth]{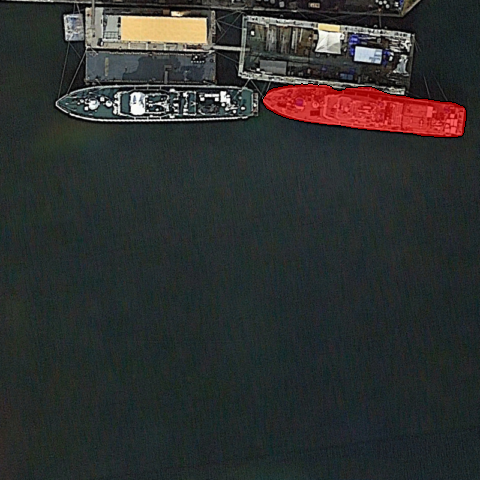}
  \end{subfigure}
  \begin{subfigure}[t]{0.155\linewidth}
  \includegraphics[width=1\linewidth]{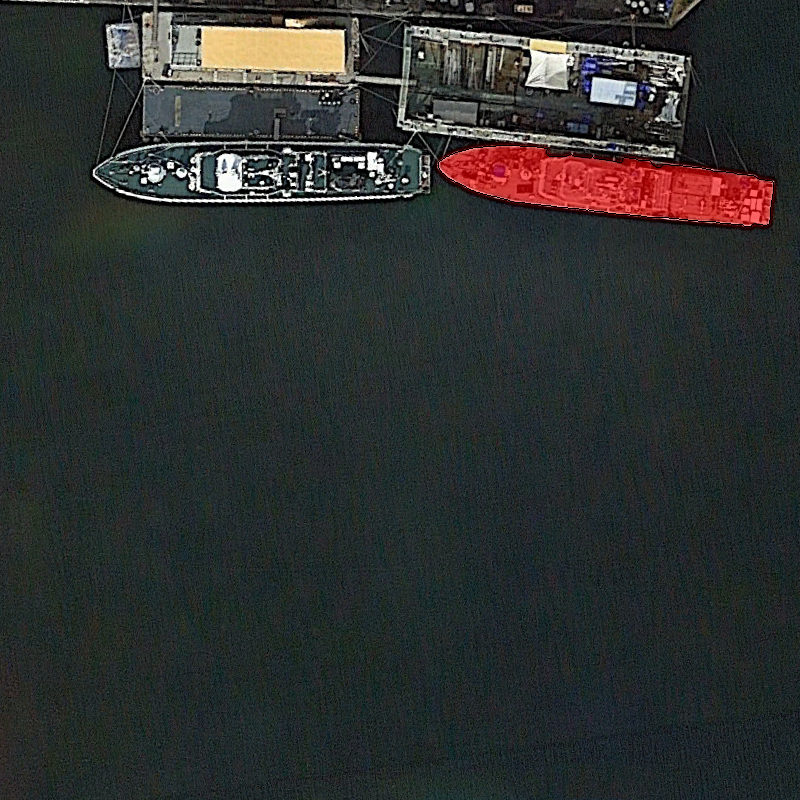}
  \end{subfigure}
  \begin{subfigure}[t]{0.155\linewidth}
  \includegraphics[width=1\linewidth]{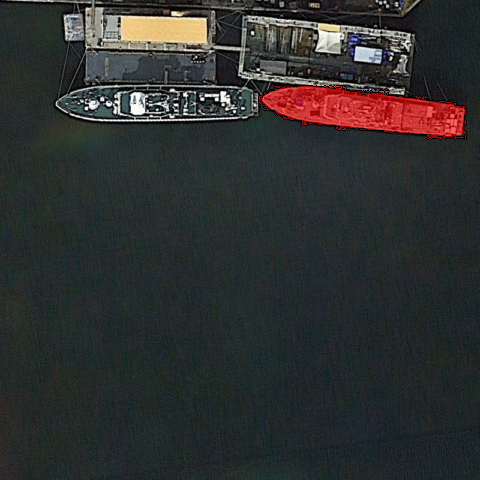}
  \end{subfigure}

  \vspace{-5mm} 
  
  \begin{subfigure}[t]{1\linewidth}
  \subcaption*{\textbf{$Expression$:} A basketball court in the middle}
  \end{subfigure}
  
  \begin{subfigure}[t]{0.155\linewidth}
  \includegraphics[width=1\linewidth]{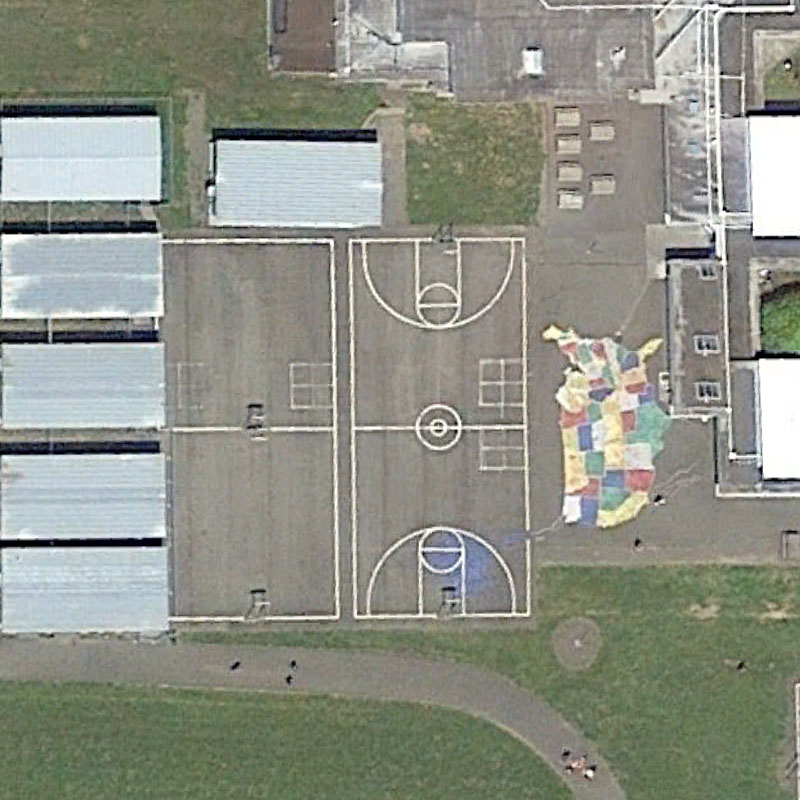}
  \end{subfigure}
  \begin{subfigure}[t]{0.155\linewidth}
  \includegraphics[width=1\linewidth]{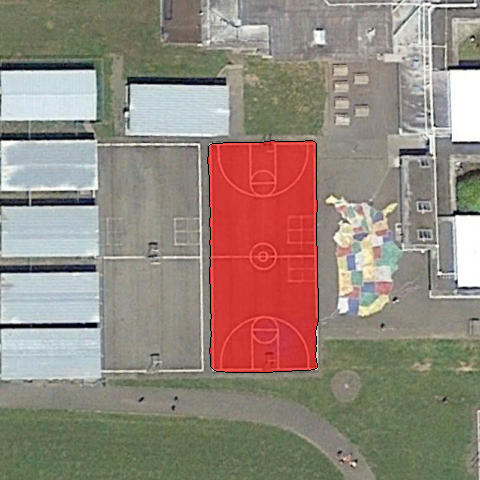}
  \end{subfigure}
  \begin{subfigure}[t]{0.155\linewidth}
  \includegraphics[width=1\linewidth]{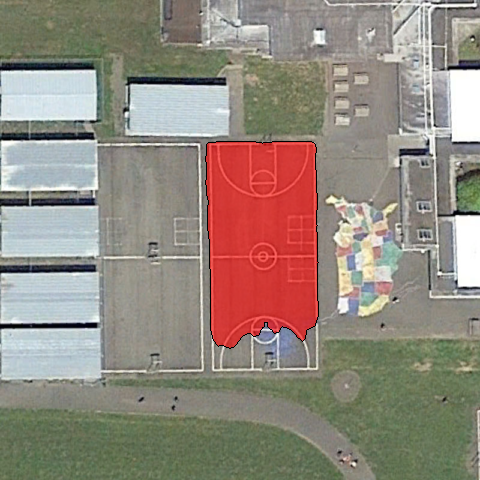}
  \end{subfigure}
  \begin{subfigure}[t]{0.155\linewidth}
  \includegraphics[width=1\linewidth]{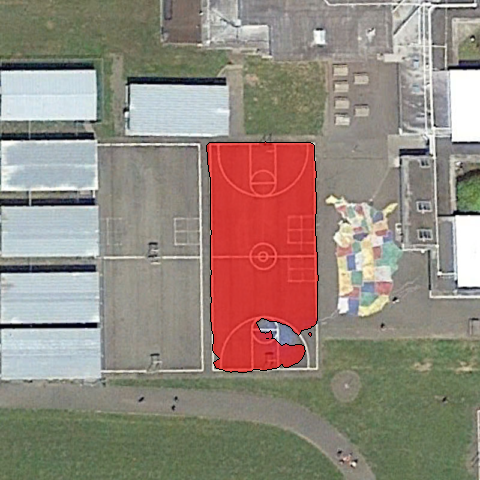}
  \end{subfigure}
  \begin{subfigure}[t]{0.155\linewidth}
  \includegraphics[width=1\linewidth]{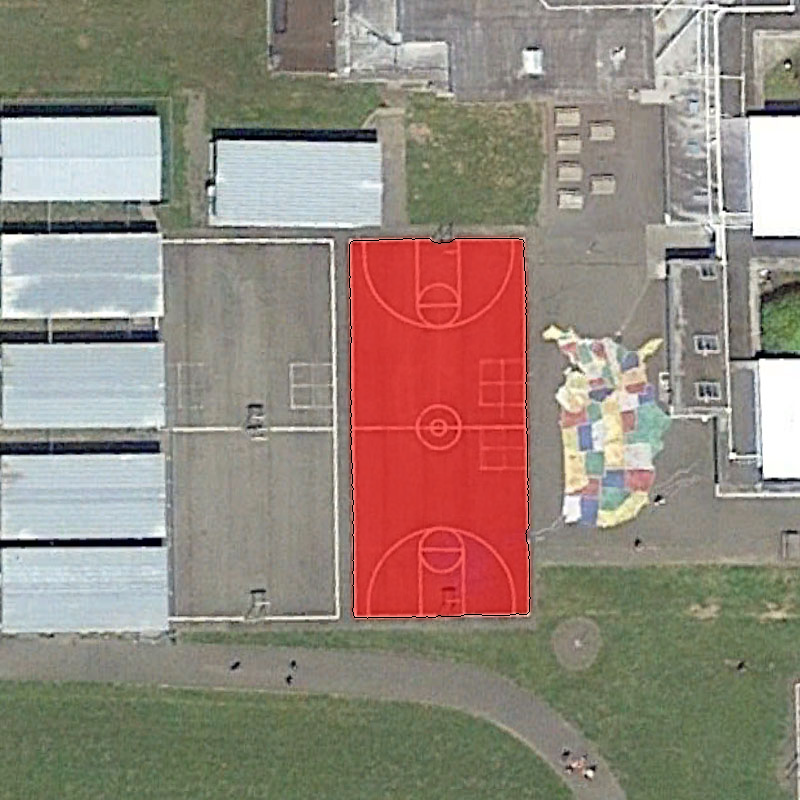}
  \end{subfigure}
  \begin{subfigure}[t]{0.155\linewidth}
  \includegraphics[width=1\linewidth]{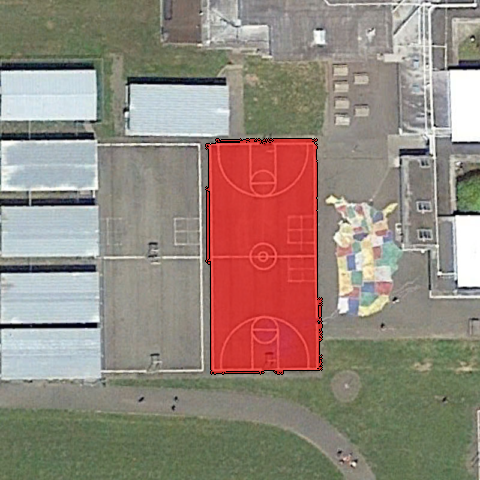}
  \end{subfigure}

  \vspace{-5mm} 
  
  \begin{subfigure}[t]{1\linewidth}
  \subcaption*{\textbf{$Expression$:} A small ground track field on the left}
  \end{subfigure}
  
  \begin{subfigure}[t]{0.155\linewidth}
  \includegraphics[width=1\linewidth]{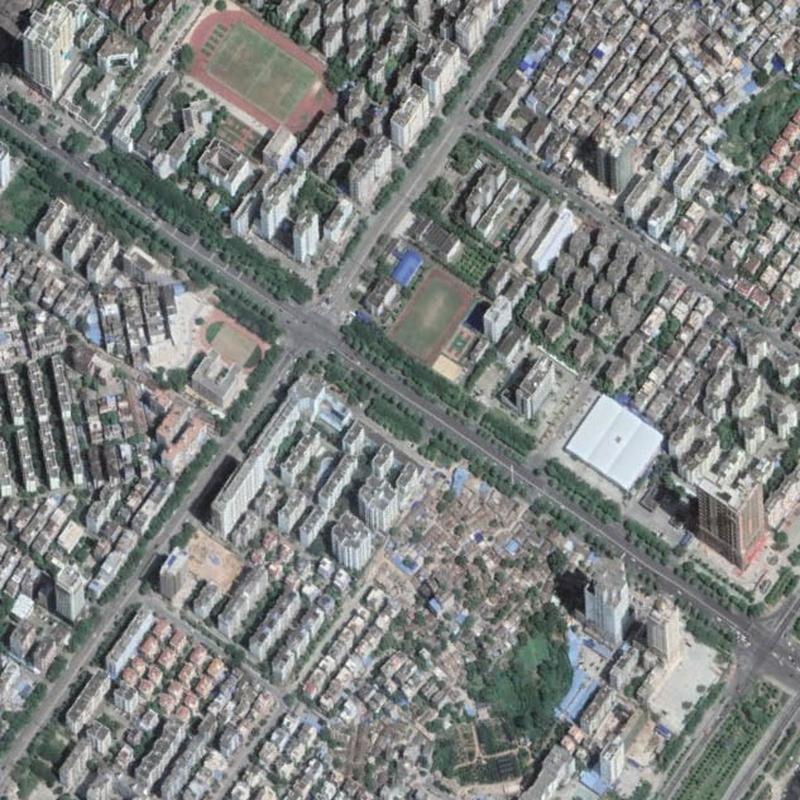}
  \end{subfigure}
  \begin{subfigure}[t]{0.155\linewidth}
  \includegraphics[width=1\linewidth]{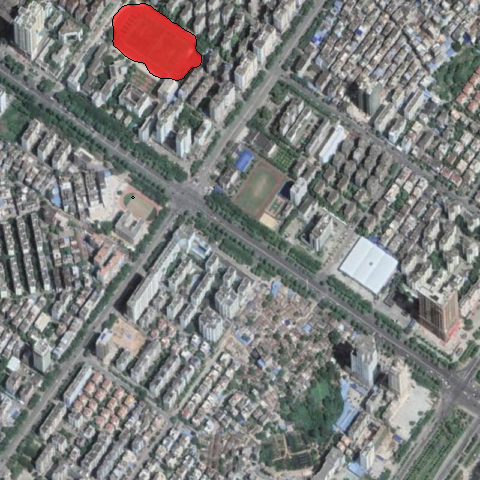}
  \end{subfigure}
  \begin{subfigure}[t]{0.155\linewidth}
  \includegraphics[width=1\linewidth]{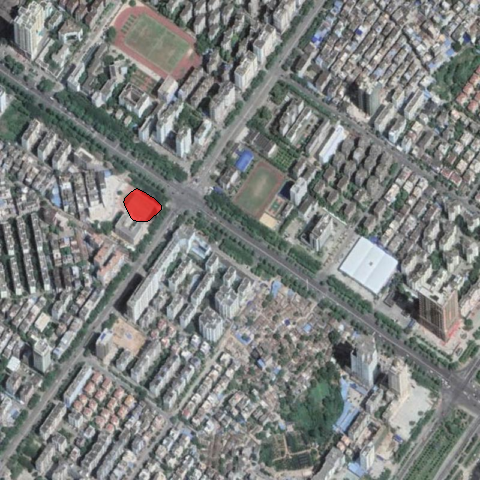}
  \end{subfigure}
  \begin{subfigure}[t]{0.155\linewidth}
  \includegraphics[width=1\linewidth]{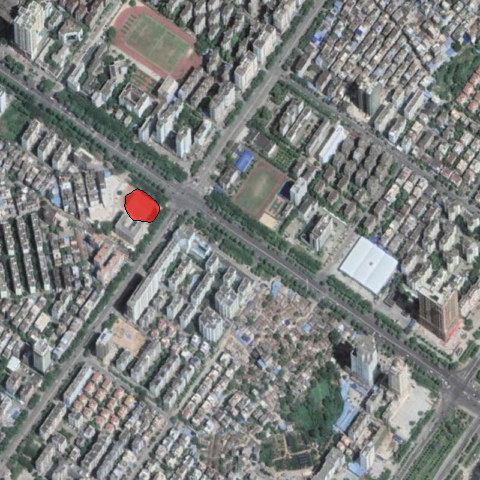}
  \end{subfigure}
  \begin{subfigure}[t]{0.155\linewidth}
  \includegraphics[width=1\linewidth]{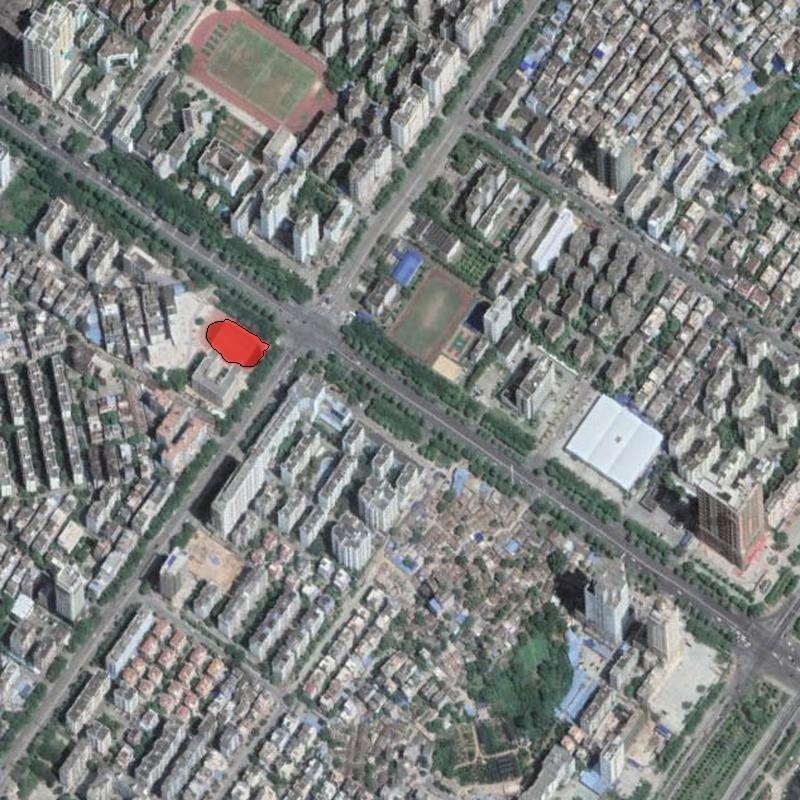}
  \end{subfigure}
  \begin{subfigure}[t]{0.155\linewidth}
  \includegraphics[width=1\linewidth]{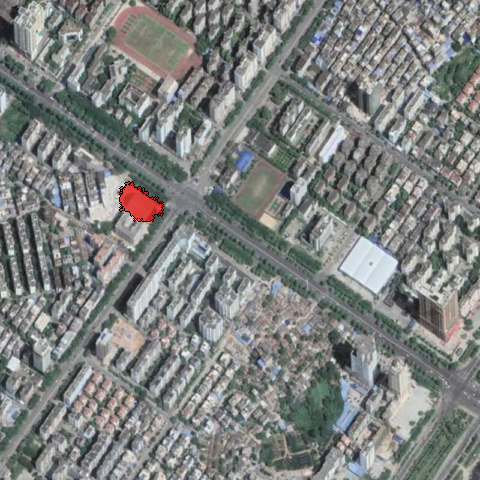}
  \end{subfigure}

  \vspace{-5mm} 

  \begin{subfigure}[t]{1\linewidth}
  \subcaption*{\textbf{$Expression$:} An airplane on the top}
  \end{subfigure}
  
  \begin{subfigure}[t]{0.155\linewidth}
  \includegraphics[width=1\linewidth]{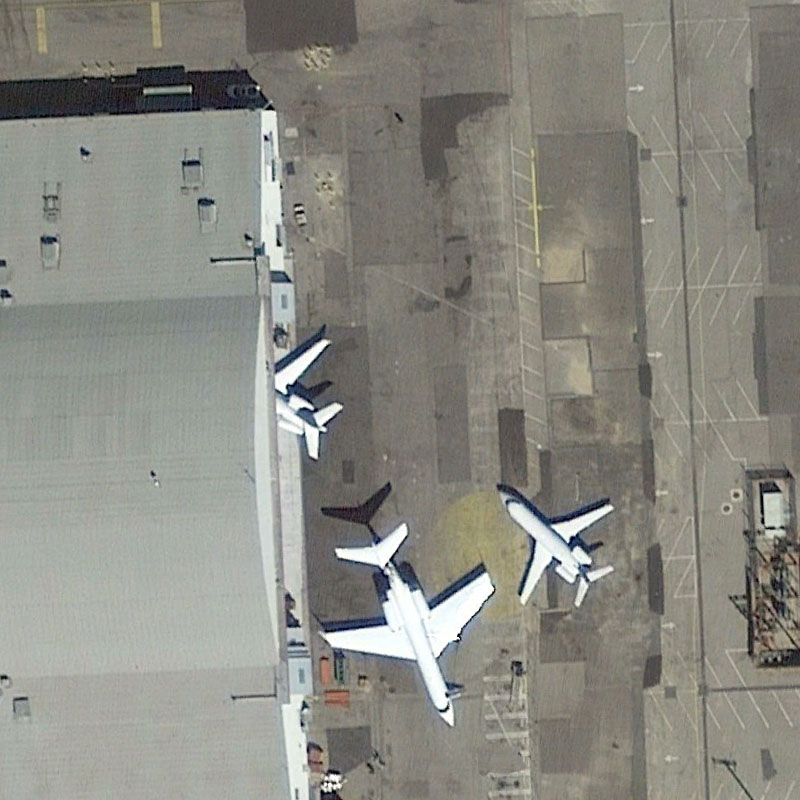}
  \end{subfigure}
  \begin{subfigure}[t]{0.155\linewidth}
  \includegraphics[width=1\linewidth]{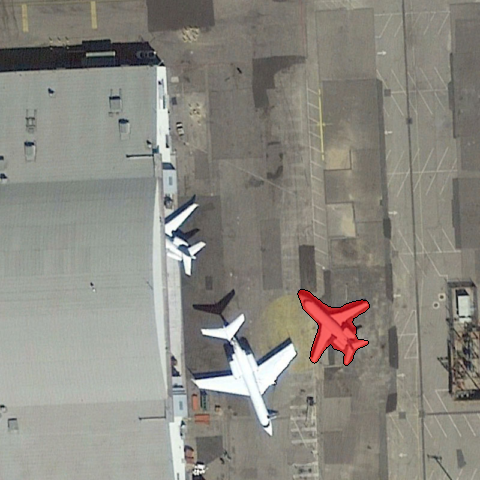}
  \end{subfigure}
  \begin{subfigure}[t]{0.155\linewidth}
  \includegraphics[width=1\linewidth]{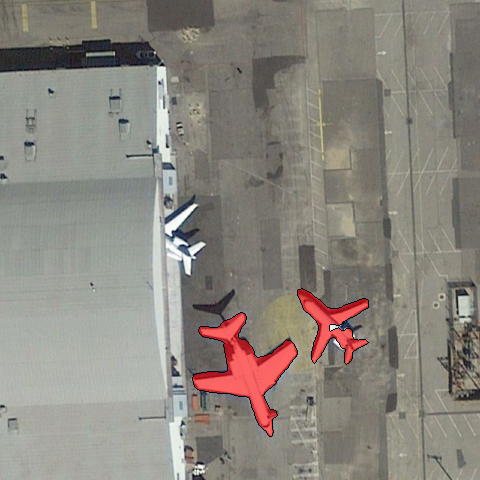}
  \end{subfigure}
  \begin{subfigure}[t]{0.155\linewidth}
  \includegraphics[width=1\linewidth]{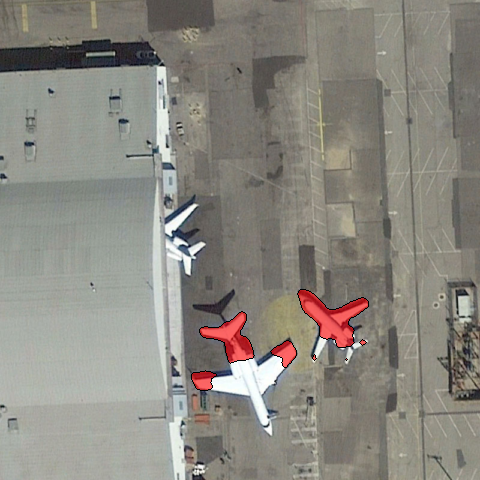}
  \end{subfigure}
  \begin{subfigure}[t]{0.155\linewidth}
  \includegraphics[width=1\linewidth]{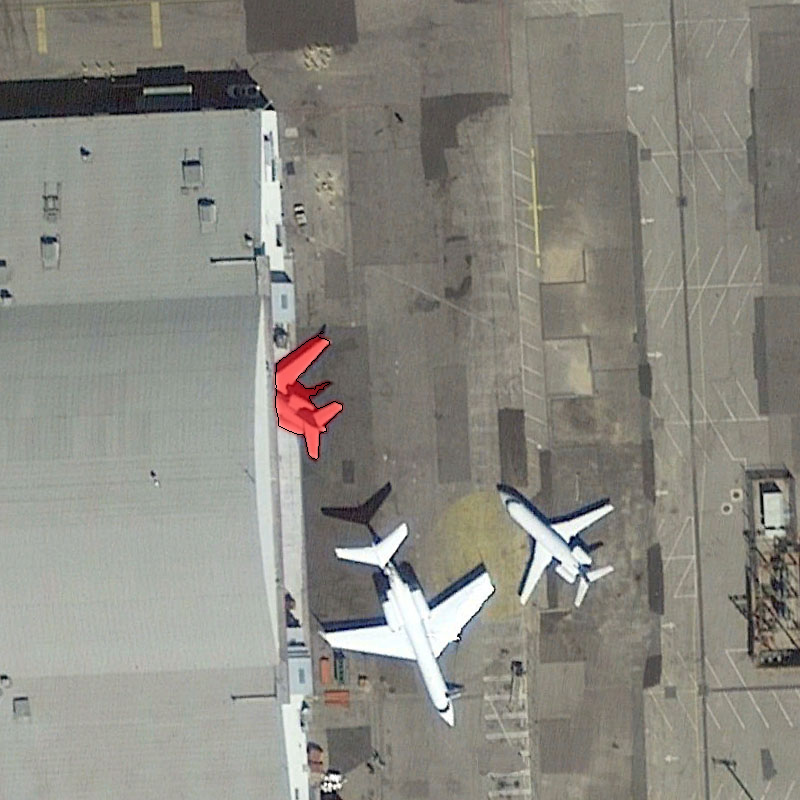}
  \end{subfigure}
  \begin{subfigure}[t]{0.155\linewidth}
  \includegraphics[width=1\linewidth]{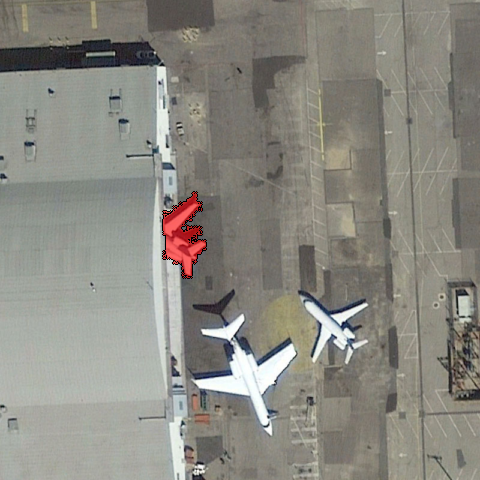}
  \end{subfigure}

  \vspace{-5mm}

  \begin{subfigure}[t]{1\linewidth}
  \subcaption*{\textbf{$Expression$:} The small ship on the far right}
  \end{subfigure}
  
  \begin{subfigure}[t]{0.155\linewidth}
  \includegraphics[width=1\linewidth]{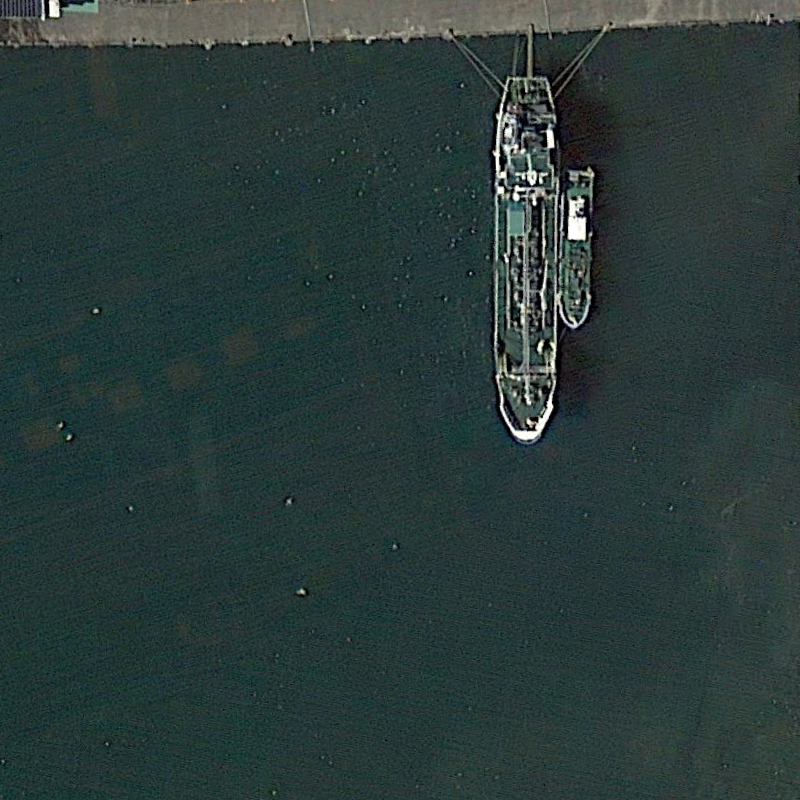}
  \subcaption*{Input Image}
  \end{subfigure}
  \begin{subfigure}[t]{0.155\linewidth}
  \includegraphics[width=1\linewidth]{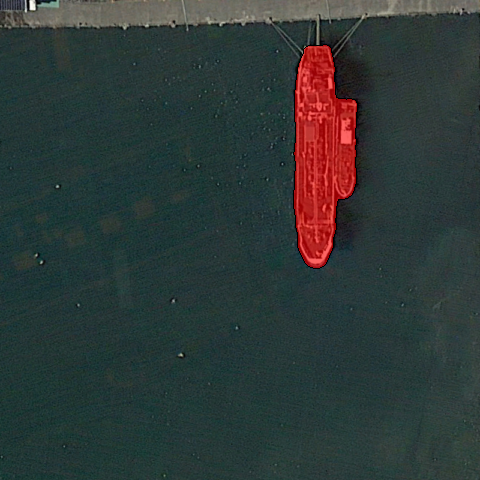}
  \subcaption*{RS2-SAM2}
  \end{subfigure}
  \begin{subfigure}[t]{0.155\linewidth}
  \includegraphics[width=1\linewidth]{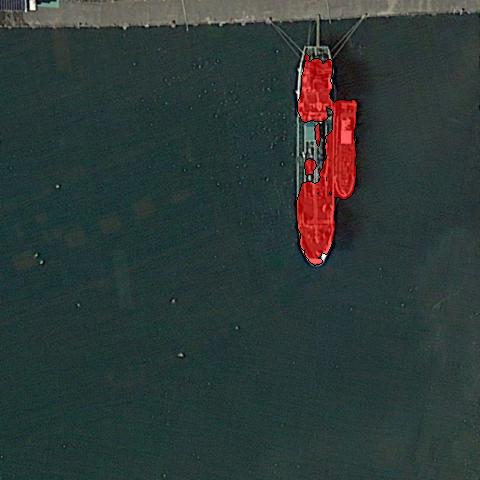}
  \subcaption*{DETRIS}
  \end{subfigure}
  \begin{subfigure}[t]{0.155\linewidth}
  \includegraphics[width=1\linewidth]{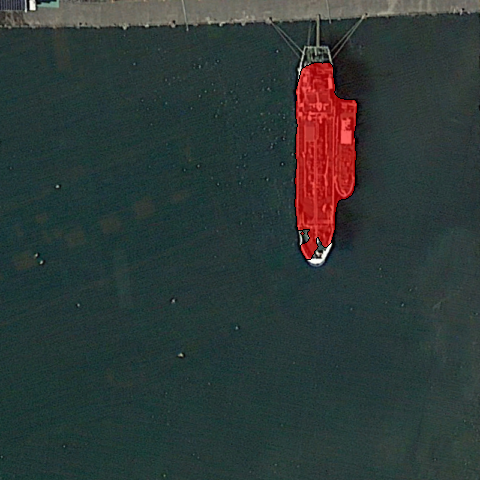}
  \subcaption*{Aurora}
  \end{subfigure}
  \begin{subfigure}[t]{0.155\linewidth}
  \includegraphics[width=1\linewidth]{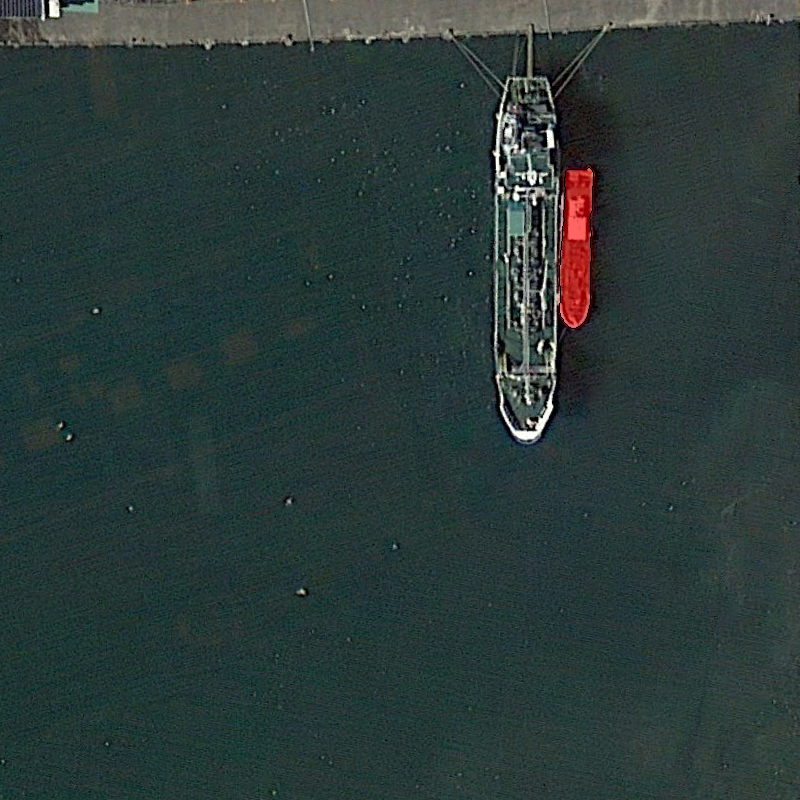}
  \subcaption*{S$^4$ECA (Ours)}
  \end{subfigure}
  \begin{subfigure}[t]{0.155\linewidth}
  \includegraphics[width=1\linewidth]{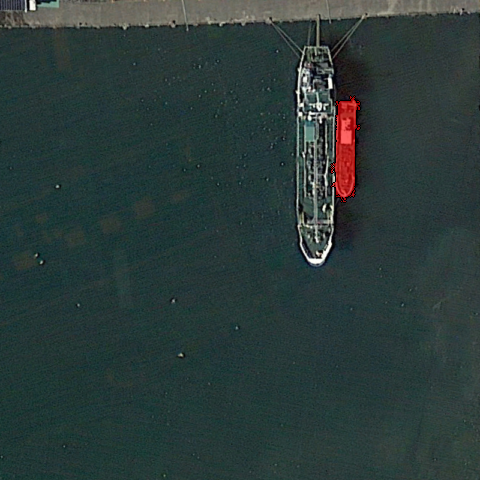}
  \subcaption*{Ground Truth}
  \end{subfigure}
  
  \vspace{-2mm}
  \caption{Qualitative comparison on the RRSIS-D \cite{liu2024rmsin} dataset.
  }
  \vspace{-5mm}
  \label{fig: rrsis results}
\end{figure*}

Figure~\ref{fig: rrsis results} presents several representative examples from the RRSIS-D dataset \cite{liu2024rmsin}.
Owing to the proposed semantic-driven textual adaptation and selective visual adaptation, S$^4$ECA consistently produced more accurate and complete segmentation masks across a wide range of object categories and scales, including chimneys, ships, sports courts, and airplanes.
Compared with existing PET methods (DETRIS and Aurora), our model more effectively produced clearer object boundaries and fewer false-positive predictions.
The advantages of S$^4$ECA became more evident in challenging scenarios involving multiple visually similar objects.
For instance, in the fifth row, several airplanes simultaneously appear in the scene, making the referring target difficult to distinguish based solely on visual saliency.
While DETRIS and Aurora were distracted by irrelevant airplanes, S$^4$ECA correctly identified the intended target by progressively propagating semantic cues through the proposed language proxies and scale-spatial adaptation mechanisms.
Compared with the recent SOTA full fine-tuning framework RS2-SAM2, S$^4$ECA  produced more complete and semantically consistent predictions despite updating only a small fraction of model parameters.
In the final example, RS2-SAM2 failed to distinguish two neighboring ships with highly similar appearances, whereas S$^4$ECA successfully captured the semantic attributes implied by the description (\textit{e.g.},~``small'' and ``right'') and localized the correct target.
This result highlights the benefit of jointly exploiting semantic, scale, and spatial priors for fine-grained cross-modal reasoning in complex aerial scenes.

\begin{figure*}[htp]
  \centering
  \begin{subfigure}[t]{1\linewidth}
  \subcaption*{\textbf{$Expression$:} van driving on the road}
  \end{subfigure}
  
  \begin{subfigure}[t]{0.155\linewidth}
  \includegraphics[width=1\linewidth]{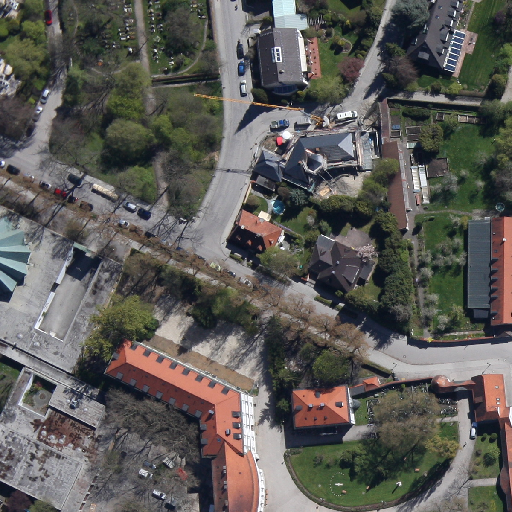}
  \end{subfigure}
  \begin{subfigure}[t]{0.155\linewidth}
  \includegraphics[width=1\linewidth]{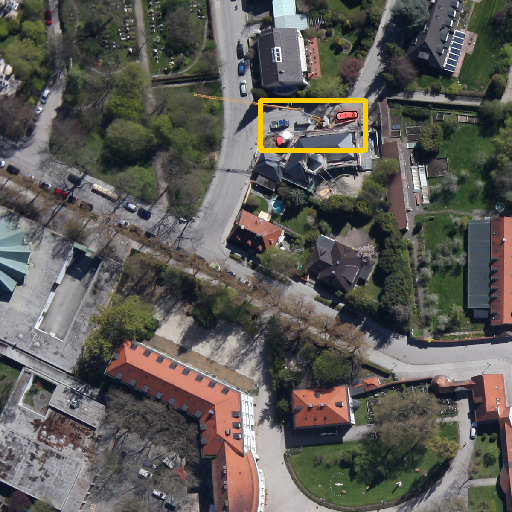}
  \end{subfigure}
  \begin{subfigure}[t]{0.155\linewidth}
  \includegraphics[width=1\linewidth]{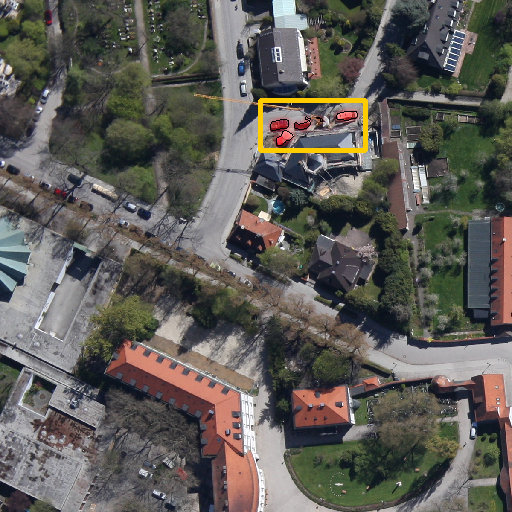}
  \end{subfigure}
  \begin{subfigure}[t]{0.155\linewidth}
  \includegraphics[width=1\linewidth]{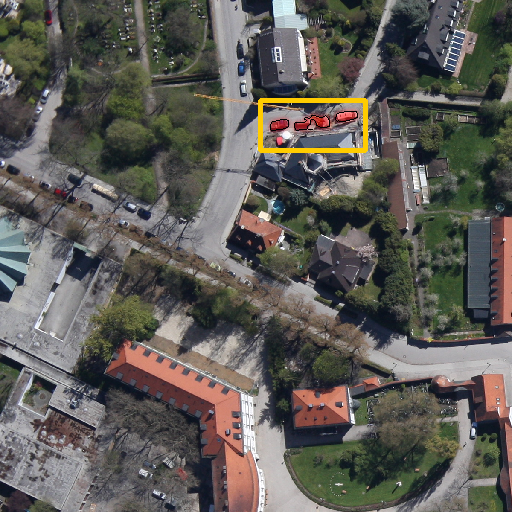}
  \end{subfigure}
  \begin{subfigure}[t]{0.155\linewidth}
  \includegraphics[width=1\linewidth]{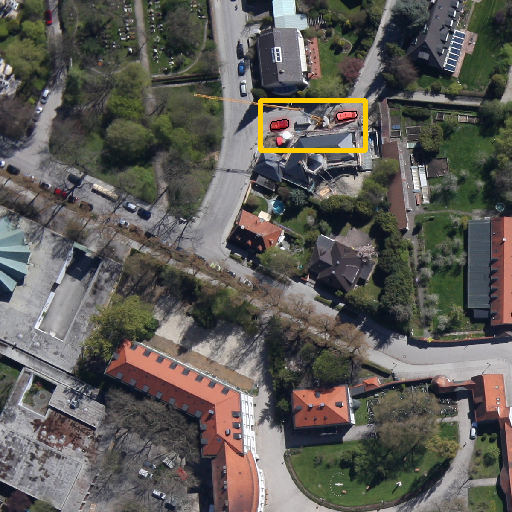}
  \end{subfigure}
  \begin{subfigure}[t]{0.155\linewidth}
  \includegraphics[width=1\linewidth]{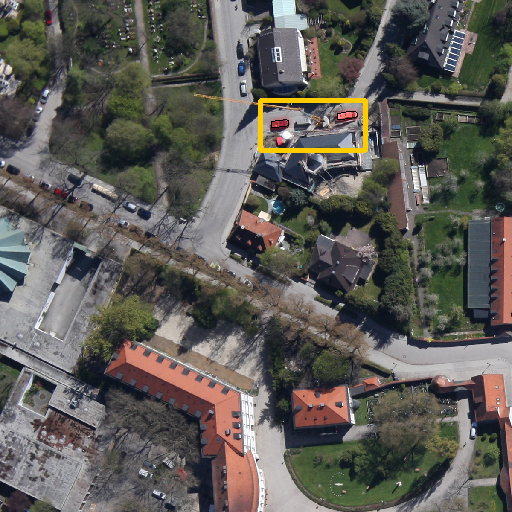}
  \end{subfigure}

  \vspace{-5mm} 
  
  \begin{subfigure}[t]{1\linewidth}
  \subcaption*{\textbf{$Expression$:} building along the road}
  \end{subfigure}
  
  \begin{subfigure}[t]{0.155\linewidth}
  \includegraphics[width=1\linewidth]{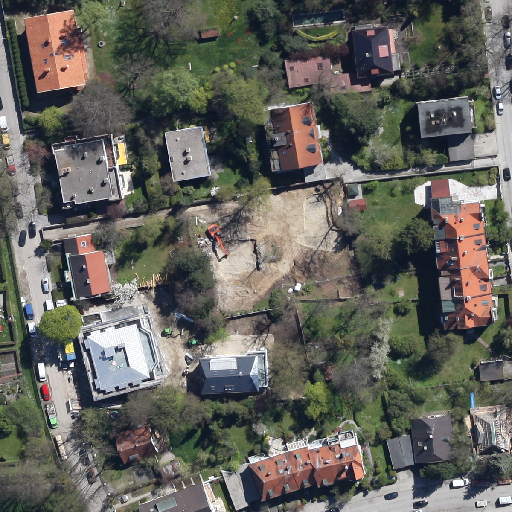}
  \end{subfigure}
  \begin{subfigure}[t]{0.155\linewidth}
  \includegraphics[width=1\linewidth]{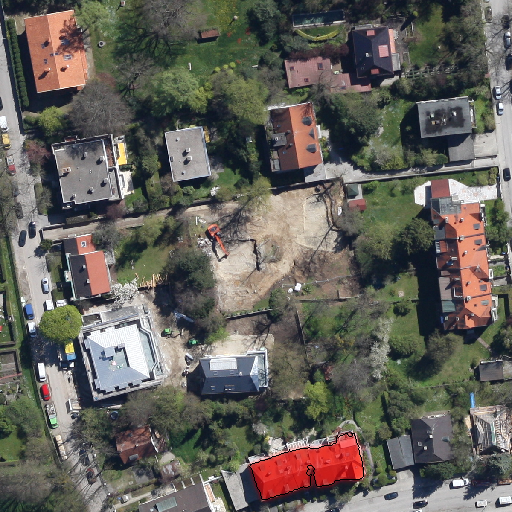}
  \end{subfigure}
  \begin{subfigure}[t]{0.155\linewidth}
  \includegraphics[width=1\linewidth]{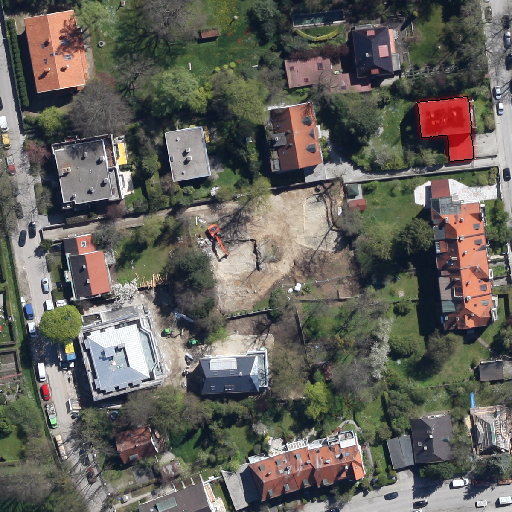}
  \end{subfigure}
  \begin{subfigure}[t]{0.155\linewidth}
  \includegraphics[width=1\linewidth]{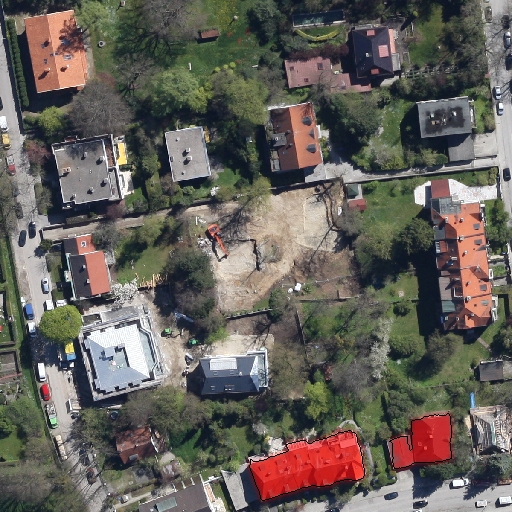}
  \end{subfigure}
  \begin{subfigure}[t]{0.155\linewidth}
  \includegraphics[width=1\linewidth]{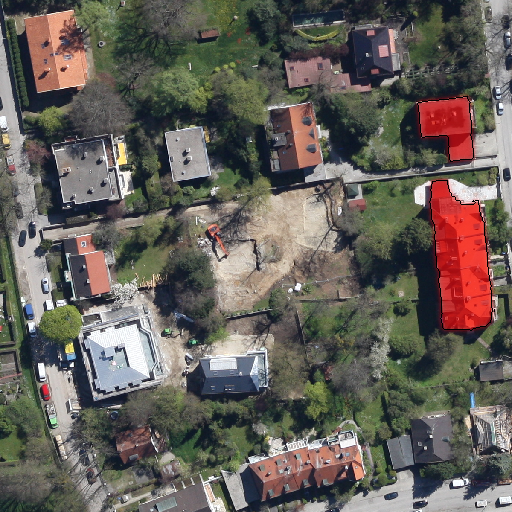}
  \end{subfigure}
  \begin{subfigure}[t]{0.155\linewidth}
  \includegraphics[width=1\linewidth]{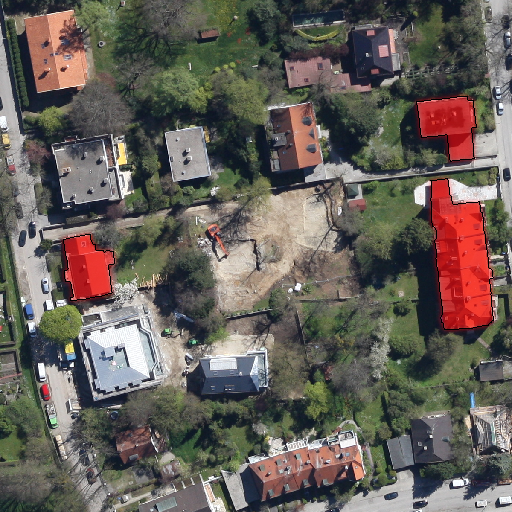}
  \end{subfigure}

  \vspace{-5mm}

  \begin{subfigure}[t]{1\linewidth}
  \subcaption*{\textbf{$Expression$:} paved road}
  \end{subfigure}
  
  \begin{subfigure}[t]{0.155\linewidth}
  \includegraphics[width=1\linewidth]{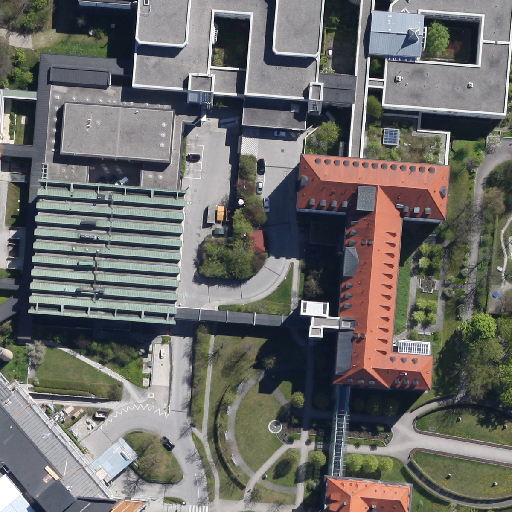}
  \subcaption*{Input Image}
  \end{subfigure}
  \begin{subfigure}[t]{0.155\linewidth}
  \includegraphics[width=1\linewidth]{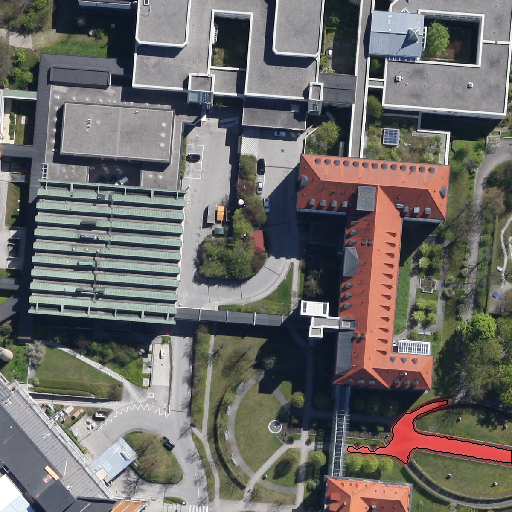}
  \subcaption*{RS2-SAM2}
  \end{subfigure}
  \begin{subfigure}[t]{0.155\linewidth}
  \includegraphics[width=1\linewidth]{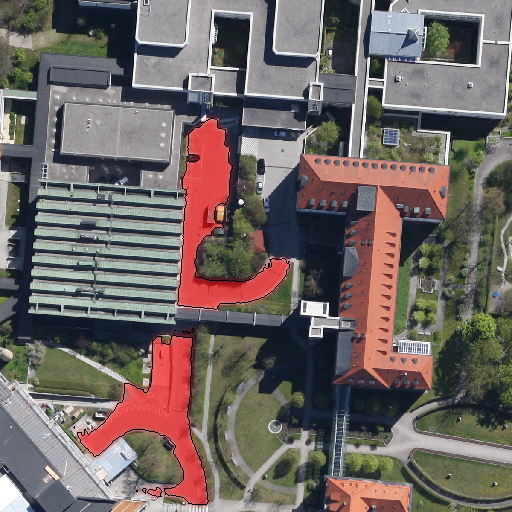}
  \subcaption*{DETRIS}
  \end{subfigure}
  \begin{subfigure}[t]{0.155\linewidth}
  \includegraphics[width=1\linewidth]{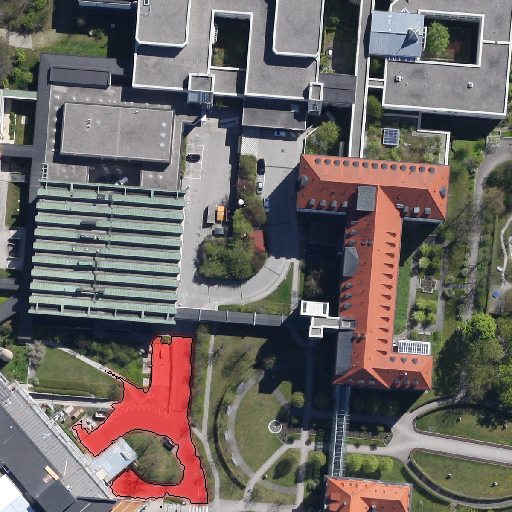}
  \subcaption*{Aurora}
  \end{subfigure}
  \begin{subfigure}[t]{0.155\linewidth}
  \includegraphics[width=1\linewidth]{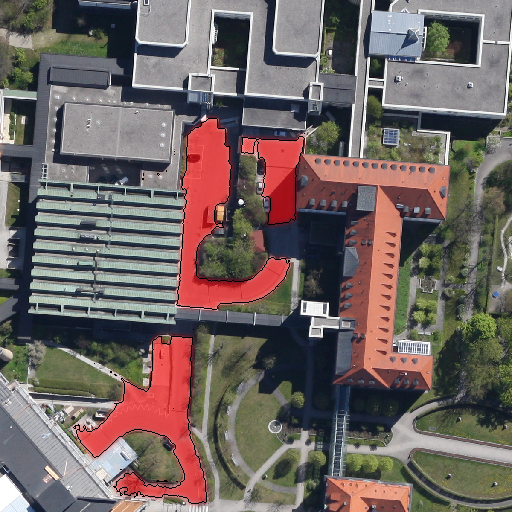}
  \subcaption*{S$^4$ECA (Ours)}
  \end{subfigure}
  \begin{subfigure}[t]{0.155\linewidth}
  \includegraphics[width=1\linewidth]{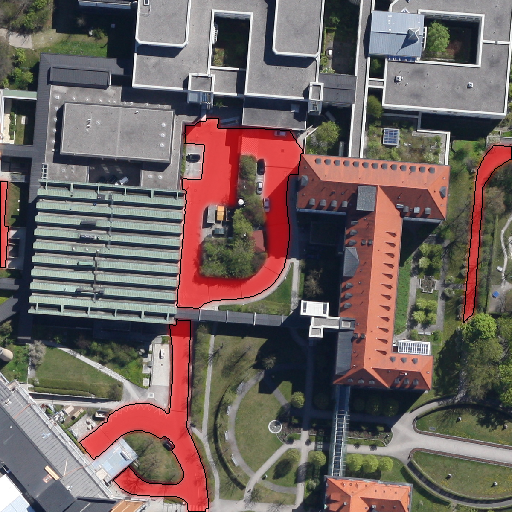}
  \subcaption*{Ground Truth}
  \end{subfigure}
  
  \vspace{-2mm}
  \caption{Qualitative comparison on the RefSegRS \cite{yuan2024rrsis} dataset.
  }
  \vspace{-5mm}
  \label{fig: refsegrs results}
\end{figure*}

Figure~\ref{fig: refsegrs results} further reports qualitative results on the RefSegRS dataset \cite{yuan2024rrsis}.
Compared with RRSIS-D, RefSegRS contains shorter and more ambiguous referring expressions, together with more complex target distributions.
Despite these challenges, S$^4$ECA still produced more precise segmentation masks than both PET-based and full fine-tuning competitors.
In particular, our model exhibited stronger robustness when the target occupies multiple regions or presents low visual contrast with surrounding backgrounds.
For instance, in the final case, only S$^4$ECA successfully identified two disconnected road segments and accurately delineated their boundaries, whereas competing methods either focused on irrelevant regions or generated incomplete masks with coarse boundaries.

Overall, these qualitative results are highly consistent with the quantitative comparisons.
They verify that the proposed dual-adapter architecture effectively enhances semantic grounding and visual feature selection, enabling more reliable cross-modal alignment and more precise target localization in complex remote sensing environments.

\subsection{Ablation Studies}
\label{sec: ablation}
To evaluate the empirical impact and architectural necessity of the sub-modules within the proposed S$^4$ECA framework, we performed a series of comprehensive ablation experiments.
Unless otherwise specified, all variants were trained and evaluated on the RRSIS-D dataset under the same experimental settings described in Section~\ref{sec: details}, with only the module under investigation being modified.


\subsubsection{Effectiveness of the Textual and Visual Adapters}
\label{abl: adapters}
To verify the contribution of each sub-module within our dual-encoder adaptation framework (\textit{i.e.},~the Semantic-driven Textual Adapter (STA) and the language-guided Scale and Spatial Selection Visual Adapter (SSVA)), we performed a progressive ablation study utilizing the DINOv2-B visual backbone.
The quantitative results are compiled in Table~\ref{tab: ablation of adapters}.
Appending the local-global feature enrichment (LGE) module to the baseline model boosted the mIoU from 61.33\% to 62.29\%, indicating that modeling local sequential dependencies helps enrich linguistic representations beyond independent token embeddings.
Building upon the LGE, the integration of the query-based semantic distillation (QSD) module further accelerated performance, raising the mIoU to 64.09\%, demonstrating the effectiveness of representing high-level semantic concepts through learnable language proxies.
Enabling early multimodal interaction via the cross-modal propagation (CMP) module pushed the text-only adapted configuration to 65.31\% mIoU, outperforming the baseline by 3.98\% while generating consistent accuracy margins across all precision metrics.
Taken together, these configurations confirm that the STA successfully captures both local and global linguistic structures and provides vital early grounding by driving semantic propagation before deep fusion occurs.
A parallel ablation trajectory is executed on the visual backbone to scrutinize our targeted scale- and spatial-selection strategies.
Incorporating the gated scale selection (GSS) module yielded a substantial improvement of 5.06\% and 3.21\% in Pr@0.5 and mIoU over the baseline, highlighting the importance of dynamically selecting informative receptive fields according to the linguistic context.
The subsequent introduction of the spatial reweighting selection (SRS) module further improved mIoU from 64.54\% to 66.25\%, suggesting that text-conditioned spatial reweighting effectively suppresses background distractions while preserving relevant target regions.
Equipping the model with the language-guided visual adaptation (LVA) module increased the mIoU to 67.92\% and drove the strict precision metric Pr@0.9 to 28.56\%, confirming the complementary roles of scale and spatial priors inside the SSVA during visual feature adaptation.
When both the STA and SSVA were jointly employed, the complete S$^4$ECA framework achieved the best performance across all evaluation metrics, reaching 78.75\% oIoU and 68.43\% mIoU.
Compared with the baseline, this corresponded to improvements of 3.06\% and 7.10\%, respectively.
The consistent gains demonstrate that the textual and visual adapters provide complementary benefits.
Specifically, the textual adapter strengthens semantic understanding and early cross-modal grounding, while the visual adapter enhances scale-aware and spatially focused feature extraction.
Together, they establish more discriminative multimodal representations and significantly improve referring segmentation performance in complex remote sensing scenes.

\begin{table}\scriptsize
  \caption{Ablation study of the proposed textual and visual adapter components in S$^4$ECA-B on the RRSIS-D test set.
  }
  \centering
  \setlength\tabcolsep{3pt}
  \begin{tabular}{p{4.8cm}ccccc}
  \toprule
  Method & Pr@0.5 & Pr@0.7 & Pr@0.9 & oIoU & mIoU\\
 \midrule
  Baseline & 70.37 & 52.15 & 23.86 & 75.69 & 61.33\\
  \midrule
  Baseline + LGE & 72.15 & 53.77 & 24.69 & 75.63 & 62.29\\
  Baseline + LGE + QSD & 74.07 & 55.45 & 25.55 & 76.28 & 64.09\\
  Baseline + LGE + QSD + CMP (STA) & 75.86 & 57.05 & 26.34 & 77.11 & 65.31\\
  \midrule
  Baseline + GSS & 75.43 & 55.26 & 25.85 & 76.33 & 64.54\\
  Baseline + GSS + SRS & 76.61 & 57.17 & 27.36 & 77.09 & 66.25\\
  Baseline + GSS + SRS + LVA (SSVA) & 77.14 & 59.29 & 28.56 & 78.20 & 67.92\\
  \midrule
  STA + SSVA (S$^4$ECA)  &  \textbf{78.28} & \textbf{61.10} & \textbf{30.03} & \textbf{78.75} & \textbf{68.43}\\
  \bottomrule
  \end{tabular}
  \label{tab: ablation of adapters}
\end{table}

\subsubsection{Analysis of Alternative Textual Adapter Designs}
\label{abl: sta}
Table~\ref{tab: ablation of sta} compares the proposed textual adapter (STA) against several representative alternatives to evaluate the effectiveness of its key designs.
We first investigated the textual feature enhancement module by replacing the proposed LGE with the dense mixture of convolutions (D-MoC) adopted in DETRIS \cite{huang2025densely}.
Although D-MoC improves textual representations through multi-branch 1D convolutions, it primarily focuses on sequential dependencies.
In contrast, LGE employs cascaded convolutions and bidirectional processing to jointly capture local and long-range linguistic dependencies, thereby compensating for the limited modeling capability of the Transformer-based CLIP text encoder.
As shown in the upper section of Table~\ref{tab: ablation of sta}, replacing LGE with D-MoC resulted in consistent performance degradation across all evaluation metrics.
These results demonstrate that jointly modeling local and global linguistic structures provides more discriminative textual representations.
We further compared the entire textual adapter with two representative textual adaptation paradigms.
The first employed an intra-modal self-attention mechanism similar to DETRIS, which updates textual features solely through token-token interactions.
The second adopted a conventional token-patch cross-attention strategy similar to Aurora \cite{yin2026aurora}, where all textual tokens directly interact with visual features.
As shown in the bottom section of Table~\ref{tab: ablation of sta}, STA consistently achieved the best results across all metrics.
Compared with the intra-modal adapter, STA improved Pr@0.5 and mIoU by 4.07\% and 1.67\%, respectively, while outperforming the cross-modal variant by 2.19\% and 0.59\%, respectively.
These results verify that both semantic-driven interaction and early cross-modal grounding are crucial for effective parameter-efficient adaptation in RRSIS.

\begin{table}\scriptsize
  \caption{Comparison of alternative textual adaptation strategies in S$^4$ECA-B on the RRSIS-D test set.
  The first group evaluates different textual feature enhancement modules, while the second group compares alternative textual adaptation mechanisms.
  }
  \centering
  \setlength\tabcolsep{3pt}
  \begin{tabular}{p{4.2cm}ccccc}
  \toprule
  Method & Pr@0.5 & Pr@0.7 & Pr@0.9 & oIoU & mIoU\\
 \midrule
  D-MoC \cite{huang2025densely} & 76.44 & 60.66 & 29.32 & 78.08 & 67.61\\
  LGE (ours) &  \textbf{78.28} & \textbf{61.10} & \textbf{30.03} & \textbf{78.75} & \textbf{68.43}\\
  \midrule
  Self-attention-based TA \cite{huang2025densely} & 74.21 & 56.34 & 26.86 & 77.18 & 66.76\\
  Cross-attention-based TA \cite{yin2026aurora} & 76.09 & 58.76 & 28.11 & 78.02 & 67.84\\
  STA (Ours)  &  \textbf{78.28} & \textbf{61.10} & \textbf{30.03} & \textbf{78.75} & \textbf{68.43}\\
  \bottomrule
  \end{tabular}
  \label{tab: ablation of sta}
\end{table}

We further investigated the influence of the number of language proxies in STA.
As shown in Table~\ref{tab: ablation of proxies}, the model performance is highly sensitive to the choice of $Z$.
When only a single proxy was employed, the model exhibited a noticeable performance degradation across all metrics.
This suggests that a single semantic representation is insufficient to capture the diverse linguistic cues contained in referring expressions, particularly in remote sensing scenarios where object categories, attributes, and spatial descriptions often coexist.
Increasing the number of proxies to 3 led to substantial performance gains, while the best overall performance was achieved with $Z=5$.
These results indicate that five language proxies provide a favorable balance between semantic diversity and representation compactness, enabling effective cross-modal interaction while avoiding redundant information.
When the number of proxies was further increased to 10, the performance began to decline on most evaluation metrics despite a marginal improvement in Pr@0.5.
Therefore, we adopt $Z=5$ in all experiments as it offers the best trade-off between representational capacity and adaptation effectiveness.

\begin{table}\scriptsize
  \caption{Influence of the number of language proxies ($Z$) in the proposed STA on the RRSIS-D test set.
  }
  \centering
  \setlength\tabcolsep{3pt}
  \begin{tabular}{p{3.2cm}ccccc}
  \toprule
  Method & Pr@0.5 & Pr@0.7 & Pr@0.9 & oIoU & mIoU\\
 \midrule
  $Z=1$ & 72.66 & 52.03 & 24.27 & 75.16 & 65.32\\
  $Z=3$  & 77.76 & \textbf{61.93} & 29.51 & 78.15 & 68.15\\
  $Z=5$  (Ours) & 78.28 & 61.10 & \textbf{30.03} & \textbf{78.75} & \textbf{68.43}\\
  $Z=10$ & \textbf{78.53} & 59.38 & 28.30 & 78.18 & 67.94\\
  \bottomrule
  \end{tabular}
  \label{tab: ablation of proxies}
\end{table}

\subsubsection{Effectiveness of the Proposed Visual Adaptation}
\label{abl: scale and spatial}
Table~\ref{tab: ablation of ssva} evaluates the effectiveness of the proposed visual adaptation design from both scale-selection and spatial-selection perspectives.
As shown in the upper section of the table, replacing the proposed GSS module with existing scale-aware aggregation alternatives consistently degraded performance.
Specifically, Dense Aligner \cite{huang2025densely} achieved an mIoU of 66.94\%, while the multi-scale aggregation strategy adopted in Aurora \cite{yin2026aurora} improved the result to 67.37\%.
In contrast, our GSS module achieved the best performance across all evaluation metrics, reaching 78.75\% oIoU and 68.43\% mIoU.
These results indicate that simply aggregating multi-scale features is insufficient for remote sensing imagery, where object sizes vary dramatically across scenes.
By conditioning scale selection on linguistic semantics, GSS enables the model to emphasize the most relevant receptive fields for the queried target, leading to more discriminative visual representations.
The lower section further investigates different spatial relevance estimation strategies.
Replacing the proposed MLP-based SRS with a non-parametric max-response strategy yielded comparable results, suggesting that both approaches can effectively identify linguistically relevant regions.
However, our proposed MLP-based design achieved the optimal overall performance.
These results demonstrate that jointly exploiting semantic, scale, and spatial cues is critical for effective visual adaptation in complex aerial scenes.

\begin{table}\scriptsize
  \caption{Ablation study of alternative visual adaptation strategies in S$^4$ECA-B on the RRSIS-D test set.
  The upper section compares different scale-aware visual adaptation mechanisms, while the lower section evaluates alternative spatial relevance estimation strategies within the proposed SRS module.
  }
  \centering
  \setlength\tabcolsep{3pt}
  \begin{tabular}{p{4.2cm}ccccc}
  \toprule
  Method & Pr@0.5 & Pr@0.7 & Pr@0.9 & oIoU & mIoU\\
 \midrule
  Dense Aligner \cite{huang2025densely} & 75.87 & 59.61 & 28.85 & 77.60 & 66.94\\
  Multi-scale Aggregation \cite{yin2026aurora} & 76.38 & 60.27 & 29.26 & 77.94 & 67.37\\
  GSS (Ours) &  \textbf{78.28} & \textbf{61.10} & \textbf{30.03} & \textbf{78.75} & \textbf{68.43}\\
  \midrule
  Max-based SRS & 78.24 & 61.01 & \textbf{30.20} & 78.74 & 68.28\\
  MLP-based SRS (Ours)  &  \textbf{78.28} & \textbf{61.10} & 30.03 & \textbf{78.75} & \textbf{68.43}\\
  \bottomrule
  \end{tabular}
  \label{tab: ablation of ssva}
\end{table}

\subsection{Efficiency Analysis}
\label{sec: efficiency}
To assess the efficiency of the proposed S$^4$ECA framework, we compared different PET methods in terms of tunable parameters, computational complexity (GFLOPs), and segmentation performance.
For a fair comparison, we preserved the original hyperparameter configurations of the baseline methods.
Where necessary, we calibrated the bottleneck dimensions of competing adapters to align their total tunable parameter counts within an equivalent order of magnitude.

\begin{table}\scriptsize
  \caption{Comparison of parameter efficiency and segmentation performance of different PET methods using DINOv2-B \cite{oquabdinov2} as the visual backbone on the RRSIS-D test set.
  Params denotes the number of tunable parameters, reported together with the proportion relative to full fine-tuning, while GFLOPs denotes the computational cost per image during inference.
  }
  \centering
  \setlength\tabcolsep{3pt}
  \begin{tabular}{p{3cm}ccccccc}
  \toprule
  Method & Params (M) & GFLOPs (G) & Pr@0.5 & Pr@0.7 & Pr@0.9 & oIoU & mIoU\\
 \midrule
  Full Fine-tuning  & 149.97 (100\%) & 168.73 & 68.32 & 50.88 & 22.83 & 74.97 & 59.16\\
  Fix Backbone & 0.00 (0.00\%) &  168.73 & 70.37 & 52.15 & 23.86 & 75.69 & 61.33\\
  \midrule
  Adapter \cite{houlsby2019adapter} & 1.98 (1.32\%) & 169.94 & 70.92 & 50.32 & 23.08 & 74.26 & 61.44\\
  Compacter \cite{karimi2021compacter} & 1.62 (1.08\%) & 169.35 & 72.43 & 53.70 & 24.42 & 76.02 & 62.86\\
  AdaptFormer \cite{chen2022adaptformer} & 1.98 (1.32\%) & 169.73 & 72.02 & 53.67 & 24.60 & 76.68 & 62.47\\
  LoRA ($r=8$) \cite{hu2022lora} & 1.57 (1.05\%) &  \textbf{169.09} & 73.45 & 54.94 & 24.95 & 76.41 & 63.55\\
  ETRIS \cite{xu2023etris} & \textbf{1.38 (0.92\%)} & 170.17 & 72.16 & 53.07 & 23.25 & 76.37 & 61.78\\
  DETRIS \cite{huang2025densely} & 2.71 (1.81\%) & 170.55 & 71.77 & 53.36 & 23.80 & 76.27 & 62.81\\ 
  
  S$^4$ECA-B (Ours) & 3.65 (2.43\%) & 169.45 &  \textbf{78.28} & \textbf{61.10} & \textbf{30.03} & \textbf{78.75} & \textbf{68.43}\\
  \bottomrule
  \end{tabular}
  \label{tab: efficiency on dino}
\end{table}

Table~\ref{tab: efficiency on dino} compares different PET methods under a DINOv2-B backbone.
Compared to the traditional full fine-tuning paradigm, all PET frameworks substantially reduced the number of tunable parameters while simultaneously elevating segmentation accuracy.
This dual advantage demonstrates that the PET paradigm effectively preserves the generalized representations embedded within pre-trained foundation models.
The superior performance of the fix backbone variant over full fine-tuning further underscores the efficacy of knowledge preservation, a benefit that PET frameworks amplify even further \cite{xu2023etris, wang2024barleria, yu2025etog}.
Notably, ETRIS achieved the smallest parameter budget with only 1.38M tunable parameters (0.92\%), while LoRA \cite{hu2022lora} required merely 1.57M parameters and introduced the lowest computational overhead among all PET variants.
Despite their efficiency, generic PET methods provided only moderate improvements over the frozen-backbone baseline.
For instance, Adapter \cite{houlsby2019adapter}, Compacter \cite{karimi2021compacter}, AdaptFormer \cite{chen2022adaptformer}, and LoRA improved mIoU by at most 2.22 percentage points compared with the fixed-backbone setting.
These results suggest that simply introducing lightweight adaptation modules is insufficient for addressing the complex cross-modal reasoning and large domain gap inherent to remote sensing imagery.
In contrast, S$^4$ECA-B achieved the best performance across all evaluation metrics, reaching 78.75\% oIoU and 68.43\% mIoU.
Compared with DETRIS, our model improved mIoU by 5.62\% while requiring only 0.94M additional tunable parameters.
More importantly, the computational complexity of S$^4$ECA-B remained comparable to existing PET approaches.
These results demonstrate that S$^4$ECA-B achieves a highly favorable trade-off between adaptation efficiency and segmentation effectiveness among existing PET strategies.

\section{Conclusion}
\label{sec: conclusion}
In this paper, we propose S$^4$ECA, a novel parameter-efficient tuning framework for referring remote sensing image segmentation.
Rather than adapting large foundation backbones through extensive fine-tuning, S$^4$ECA adopts a dual-encoder adaptation strategy that enables efficient cross-modal reasoning while preserving the rich knowledge embedded in pre-trained models.
Specifically, we develop a semantic-driven textual adapter that exploits learnable language proxies to capture high-level linguistic semantics and establish early visual grounding during textual adaptation.
For the visual branch, we introduce a language-guided scale and spatial selection adapter that dynamically identifies informative receptive fields and emphasizes target-relevant regions, thereby reducing feature redundancy and improving cross-modal alignment.
Together, these designs enable the model to perform selective and efficient multimodal interaction throughout the feature encoding process.
Extensive experiments on the challenging RRSIS-D and RefSegRS benchmarks demonstrate that S$^4$ECA achieves state-of-the-art performance among parameter-efficient methods while updating only a small fraction of the backbone parameters.
Overall, S$^4$ECA establishes a strong and scalable framework for efficient remote sensing vision-language learning, providing a favorable balance between adaptation efficiency and segmentation accuracy.

{
\small
\bibliographystyle{plainnat}
\bibliography{refs}
}







\end{document}